%% file: tpami.tex
\documentclass[10pt,journal,compsoc,twoside]{IEEEtran}
\usepackage{booktabs}
\usepackage{algorithm}
\usepackage{setspace}
\usepackage{amssymb}
\usepackage{multirow}
\usepackage{url}
\usepackage{soul}	
\usepackage{amsmath}
\usepackage{subfigure}
\usepackage{cite}

\usepackage{tabulary}
\usepackage{tabularx}
\usepackage{tabu}

\usepackage[capitalise]{cleveref}

\ifCLASSINFOpdf
  \usepackage[pdftex]{graphicx}
\else
\fi
\usepackage{amsmath}
\usepackage{algorithmic}
\begin{document}
%
\title{MgSvF: Multi-Grained Slow vs. Fast Framework for Few-Shot Class-Incremental Learning}

\markboth{IEEE TRANSACTIONS ON PATTERN ANALYSIS AND MACHINE INTELLIGENCE, ~Vol.~XX, No.~X, March 2021}{Zhao \MakeLowercase{\textit{et al.}}: MgSvF: Multi-Grained Slow vs. Fast Framework for Few-Shot Class-Incremental Learning}
%
%
%
%

\author{Hanbin~Zhao, 
        Yongjian~Fu, 
        Mintong~Kang,
        Qi Tian,
        Fei Wu,
        and Xi~Li 
\IEEEcompsocitemizethanks{\IEEEcompsocthanksitem H. Zhao, Y. Fu, M. Kang, F. Wu, and X.~Li are with College of Computer Science and Technology, Zhejiang University, Hangzhou 310027, China. 
	\protect
	
  E-mail: \{zhaohanbin,~yjfu,~kangmintong, wufei, xilizju\}@zju.edu.cn. \protect
  \IEEEcompsocthanksitem Q.~Tian is with Cloud BU, Huawei Technologies, Shenzhen 518129, China. 
  \protect
  
  E-mail: tian.qi1@huawei.com. \protect
}
\thanks{(Corresponding author: Xi Li.)}} 

\input{sections_for_review/abstract}

\maketitle

\IEEEdisplaynontitleabstractindextext

%
\IEEEpeerreviewmaketitle

\input{sections_for_review/intro}

\input{sections_for_review/related}

\input{sections_for_review/methods}

\input{sections_for_review/eval}
\input{sections_for_review/conclusion}
  \section*{Acknowledgment}
The authors would like to thank Xuewei Li, Songyuan Li and Hui Wang for their valuable comments and suggestions.


\ifCLASSOPTIONcaptionsoff
  \newpage
\fi



%
\bibliographystyle{IEEEtran}
\bibliography{bibli}
%
%

%

%




\end{document}

%% file: sections_for_review/abstract.tex
\IEEEtitleabstractindextext{%
\begin{abstract}
As a challenging problem, few-shot class-incremental learning (FSCIL) continually learns a sequence of tasks, confronting the dilemma between slow forgetting of old knowledge and fast adaptation to new knowledge.
In this paper, we concentrate on this ``slow vs. fast'' (SvF) dilemma to determine which knowledge components to be updated in a slow fashion or a fast fashion, and thereby balance old-knowledge preservation and new-knowledge adaptation.
We propose a multi-grained SvF learning strategy to cope with the SvF dilemma from two different grains: intra-space (within the same feature space) and inter-space (between two different feature spaces). 
The proposed strategy designs a novel frequency-aware regularization to boost the intra-space SvF capability, and meanwhile develops a new feature space composition operation to enhance the inter-space SvF learning performance. With the multi-grained SvF learning strategy, our method outperforms the state-of-the-art approaches by a large margin.
\end{abstract}

\begin{IEEEkeywords}
Few-shot class-incremental learning, multi-grained, class-incremental learning
\end{IEEEkeywords}}

%% file: sections_for_review/intro.tex
\begin{figure*}[t]
	\centering
	\subfigure{}{
		\begin{minipage}[ht]{0.85\textwidth}
			\includegraphics[width = 1\columnwidth]{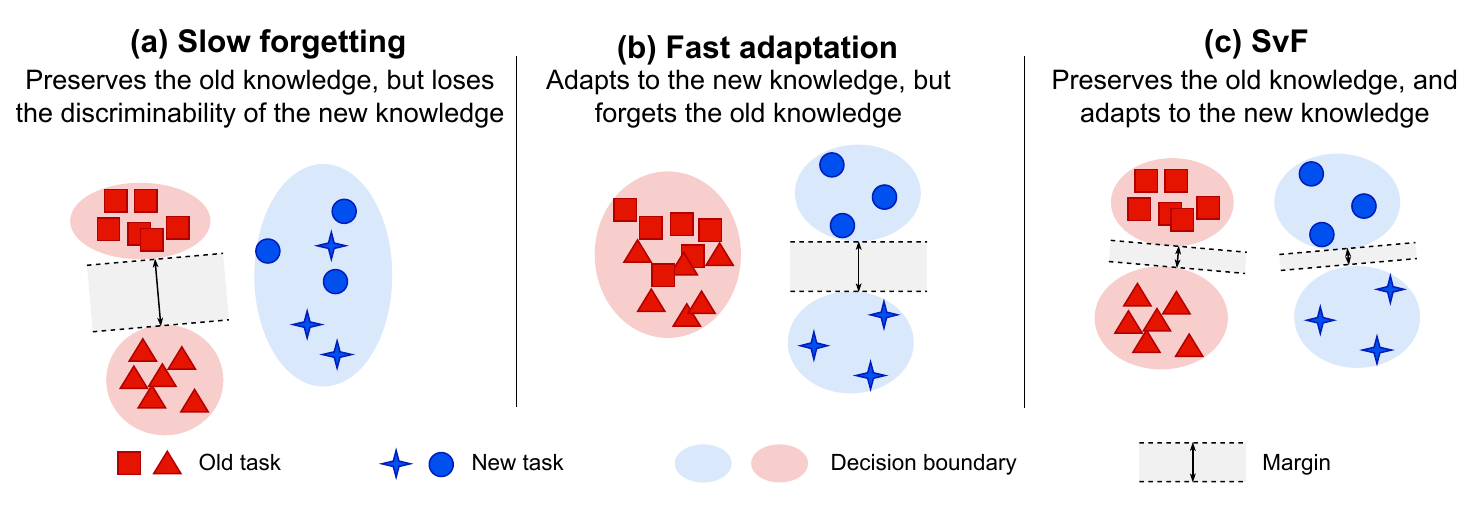}
		\end{minipage}	
	}
	\caption{Illustration of slow vs. fast analysis for few-shot class-incremental learning. (a) mainly pays attention to slow forgetting. Samples of old tasks are separated by a large margin but that of new tasks are mixed up. (b) puts emphasis on fast adaptation. New-task samples are separable while old-task samples are mixed up. (c) keeps a trade-off between slow-forgetting and fast-adaptation and solves all tasks well.}
	\label{fig:FSCIL}
\end{figure*}

\IEEEraisesectionheading{\section{Introduction}\label{sec:intro}}
\IEEEPARstart{R}{e}cent years have witnessed a great development of class-incremental learning~\cite{yu2020semantic,tao2020topology,iscen2020memory,douillard2020small,zhao2020maintaining,rajasegaran2020itaml,hayes2019remind,masana2020class}, which aims at enabling a learner to acquire new knowledge from new data while preserving the learned knowledge from previous data. 
In practice, the new knowledge from new data is often represented in a challenging few-shot learning scenario (i.e., few annotated samples), leading to a problem named few-shot class-incremental learning~\cite{tao2020few} (FSCIL).
FSCIL typically involves the learning stages of the base task (i.e., the first task with large-scale training samples) and the new tasks (with limited samples). 
In principle, FSCIL is in a dilemma between slow forgetting of old knowledge and fast adaptation to new knowledge.
As shown in Figure~\ref{fig:FSCIL}~(a) and (b), slow forgetting typically leads to underfitting on new tasks, while fast adaptation incurs a catastrophic forgetting problem.
Hence, a ``slow vs. fast'' (SvF) learning pipeline is needed to be implemented to determine which knowledge components to be updated in a slow fashion or a fast fashion, keeping a trade-off between slow-forgetting and fast-adaptation as shown in Figure~\ref{fig:FSCIL}~(c).
In this paper, we focus on investigating the SvF learning performance differences from two
different grains: within the same feature space (called intra-space SvF analysis) and between two different feature spaces (called inter-space SvF analysis).

\begin{figure}[t]
	\centering
	\subfigure{}{
		\begin{minipage}[ht]{0.499\textwidth}
			\includegraphics[width = 0.49\columnwidth]{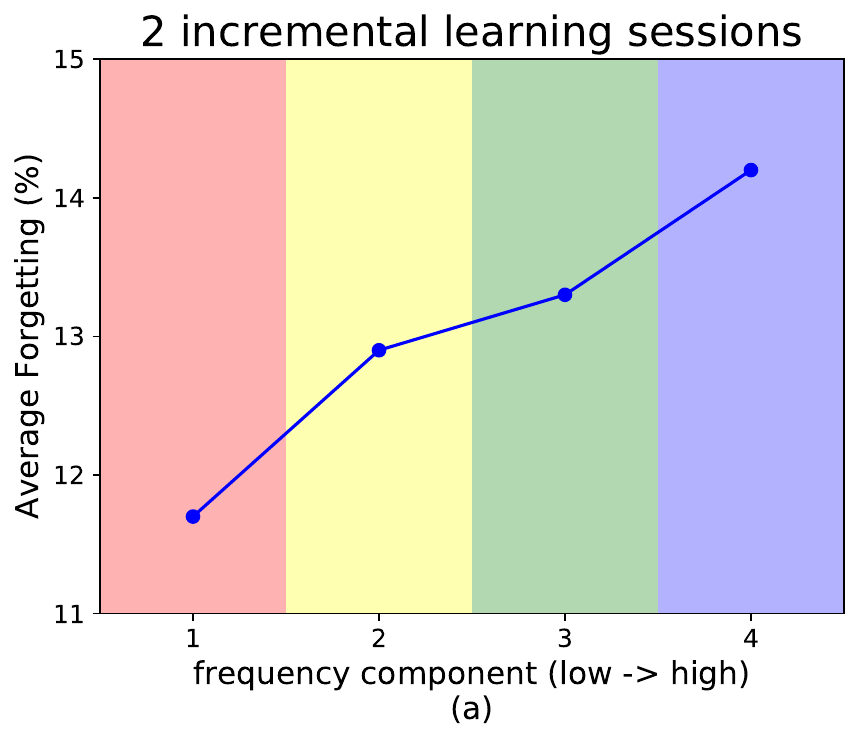}
			\includegraphics[width = 0.49\columnwidth]{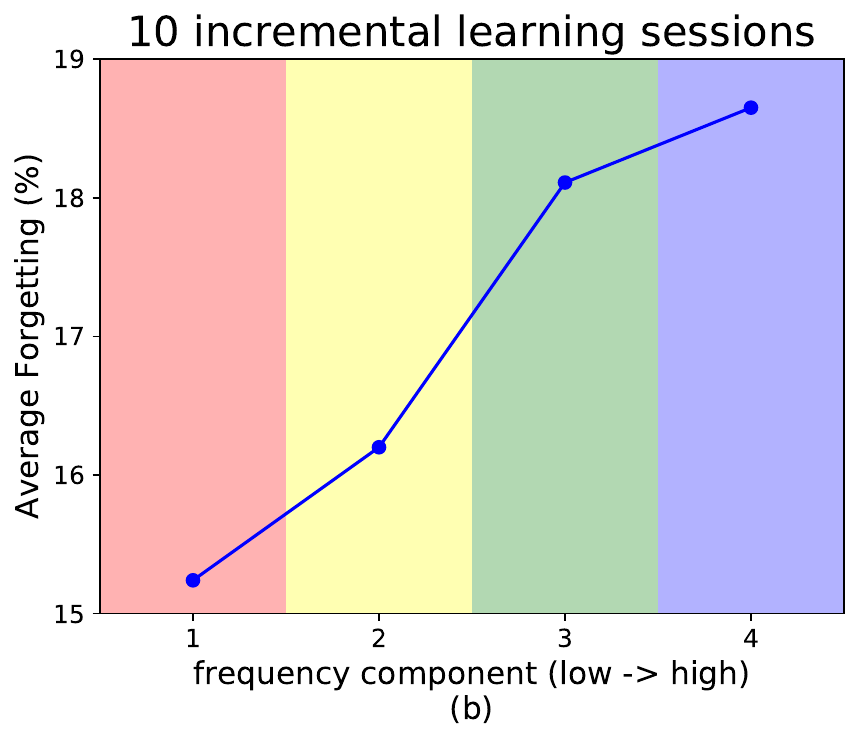}
		\end{minipage}	
	}
	\caption{Analysis of intra-space SvF on CIFAR100.
		The forgetting of previous tasks is estimated with \textit{average forgetting}~\cite{yu2020semantic,chaudhry2018riemannian}. 
		(a) Results with 2 learning sessions. (b) Results with 10 learning sessions. 
		It can be seen that different frequency components appear different characteristics for old-knowledge transfer.
		In both learning settings, the lower frequency components achieve less forgetting, and the average forgetting increases along with the frequency.
	}
	\label{fig:frequency}
\end{figure}

In the literature, a number of approaches only maintain a unified feature space for SvF learning w.r.t. different feature dimensions~\cite{tao2020few,french1999catastrophic,goodfellow2013empirical,mccloskey1989catastrophic,pfulb2019comprehensive,rebuffi2017icarl,castro2018end,hou2019learning}.
Since the unified feature space has mutually correlated feature dimensions,
it is difficult to disentangle the feature for SvF analysis. Besides, 
the learning directions for old-knowledge preservation and new-knowledge adaptation are usually 
inconsistent with each other (even contradictory sometimes). In the case of FSCIL,  
the unified feature space tends to fit the data of new-task
well, but suffers from the degradation of discriminability and generalization ability, as well as catastrophic forgetting.

Motivated by the above observations, we build an intra-space SvF feature disentanglement scheme by Discrete Cosine Transform (DCT), resulting in an orthogonal frequency space with mutually uncorrelated frequency components. 
Subsequently, we propose to evaluate the SvF performance differences w.r.t. different frequency components. 
The evaluation results indicate that different frequency components indeed appear to have different characteristics for knowledge transfer.
As shown in Figure~\ref{fig:frequency}, the low-frequency components contribute more to preserving old knowledge, and the average forgetting increases along with the frequency.
Therefore, we turn out to build the new feature space updated by frequency-aware knowledge distillation, which enforces higher weights on the regularization term of approximating the low-frequency components of the old feature space. 
Experiments show that this simple frequency-aware learning strategy results in an effective performance improvement.

Given the aforementioned frequency space, we also maintain another separate feature space that focuses on preserving the old knowledge to further enhance the discriminability. In this way, we set up an inter-space SvF learning scheme to make the separate feature space updated more slowly than the other.
In the inter-space SvF learning scheme, we propose a feature space composition operation to compose the above two spaces.
In principle, different composition operations are flexible to be used.
From extensive experiments, we find out that even an extremely simple uniform concatenation strategy can result in a dramatic performance improvement.

Overall, the main contributions of this work are summarized in the following three aspects:
1) We propose a multi-grained ``slow vs. fast'' (SvF)  learning framework for FSCIL, which aims to
balance old-knowledge preserving and new-knowledge adaptation. To our knowledge, it is the first work to introduce
multi-grained SvF into FSCIL.
2) We present two simple yet effective SvF learning strategies for intra-space and inter-space cases, that is, frequency-aware regularization and feature space composition operation.
3) Extensive experiments over all the datasets demonstrate the effectiveness of our approach as our method outperforms state-of-the-art approaches by a large margin.

%% file: sections_for_review/related.tex
\section{Related Work}
\label{sec:related}
\subsection{Incremental Learning.}
Recently, there has been a large body of research in incremental learning~\cite{de2019continual,parisi2019continual,li2020compositional,lee2020neural,adel2019continual,von2019continual,kurlecontinual,titsias2019functional, ebrahimi2019uncertainty, zeng2019continual,kundu2020class,ostapenko2019learning,aljundi2019task,lee2020continual,he2020incremental,ebrahimi2020adversarial,liumore,caccia2020online,nguyen2017variational, chen2020incremental, dang2020class,shim2020online,liu2020incdet,li2020continual,li2011graph,chen2014ranking,jiang2019learning}.
These works can be categorized into three major families: 1) architectural strategies, 2) rehearsal strategies, 3) regularization strategies.
Architectural strategies~\cite{yoon2017lifelong,li2019learn,hung2019compacting,mallya2018packnet,mallya2018piggyback,serra2018overcoming,rusu2016progressive,aljundi2017expert,rajasegaran2019random,abati2020conditional} keep the learned knowledge from previous tasks and acquire new knowledge from the current task by manipulating the network architecture, e.g., parameter masking, network pruning.
Rehearsal strategies~\cite{chaudhry2019agem,lopez2017gradient,shin2017continual,aljundi2019online,zhai2019lifelong,wu2018memory,rebuffi2017icarl,castro2018end,he2018exemplar,hou2019learning,wu2019large,liu2020mnemonics,summaira2021recent} replay old tasks information when learning the new task, and the past knowledge is memorized by storing old tasks' exemplars or old tasks data distribution via generative models. Regularization strategies~\cite{lee2017overcoming,li2017learning,nguyen2017variational,ritter2018online,kirkpatrick2017overcoming,zenke2017continual,liu2018rotate,aljundi2018memory,dhar2019learning,mirzadeh2020understanding} alleviate forgetting by regularization loss terms enabling the updated parameters of networks to retain past knowledge. 
Incremental learning is usually conducted under the task-incremental~\cite{kanakis2020reparameterizing,rosenfeld2018incremental} or the class-incremental learning scenarios~\cite{tao2020topology,perez2020incremental,cermelli2020modeling,belouadah2018deesil,xiang2019incremental,dhar2019learning}. This paper considers the latter where the task identity is non-available at inference time. Few-shot class-incremental learning~\cite{tao2020few} is a more practical and challenging problem, where only a few number of samples for new tasks are available. The aforementioned approaches resort to one unified feature space for SvF learning to balance old-task knowledge preserving and new-task knowledge adaptation. In this paper, we investigate the SvF learning performance from intra-space and inter-space grains and propose a multi-grained SvF learning strategy for FSCIL (i.e., frequency-aware regularization and feature space composition operation).

\begin{figure*}[!ht]
	\centering
	\begin{minipage}[ht]{1\textwidth}
		\includegraphics[width = 1\columnwidth]{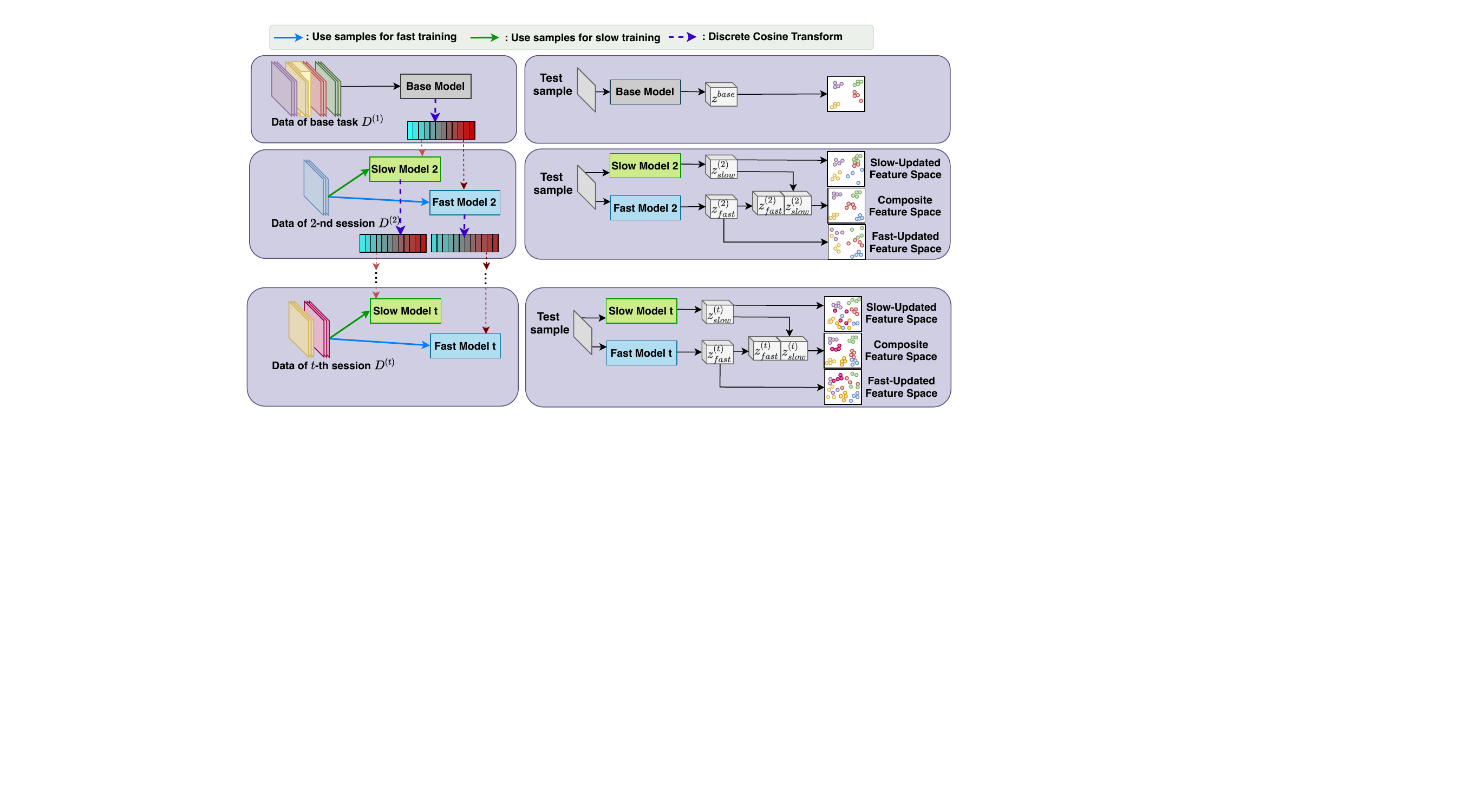}	
	\end{minipage}
	\caption{Illustration of our method. The \emph{dash lines} show the frequency-aware intra-space regularization, and the \emph{solid lines} indicate the inter-space composition operation. At the $1$-st session, a base embedding model is initially trained on a large-scale training set of base task. At the $t$-th ($t \textgreater 1$) learning session, two embedding models are fast or slowly updated on data of $t$-th task by intra-space SvF learning, then we composite the slow-updated feature space and the fast-updated feature space, and finally use the composite feature space for classification.}
	\label{fig:our_scheme}
	
\end{figure*}

\begin{table*}[!ht]
	\centering
	\caption{Main notations and symbols used throughout the paper.}
	\resizebox{0.87\textwidth}{!}{
		\begin{tabular}{c l l}
			\toprule
			\textbf{Notation} & \multicolumn{2}{c}{\textbf{Definition}} \\
			\midrule
			$D^{(t)}$&  \multicolumn{2}{l}{the training set for the $t$-th task and contains only a few samples}\\
			
			$D^{(1)}$&  \multicolumn{2}{l}{the first task (termed as a base task) and has a large number of samples}\\
			
			$C^{(t)}$&  \multicolumn{2}{l}{the set of classes of the $t$-th task}\\
			
			${\rm{dist}}(\cdot ,\cdot)$&  \multicolumn{2}{l}{the distance metric (e.g., Euclidean distance)}\\	
			
			$\mathbf{z}_j$&  \multicolumn{2}{l}{the embedding of a given sample $x_j$ in the original entangled feature space}\\	
			
			$\mathbf{u}_c$&  \multicolumn{2}{l}{the mean of embedding of class $c$}\\	
			
			$\hat{y}_j$&  \multicolumn{2}{l}{the prediction result of a given sample $x_j$}\\
			
			$T(\cdot)$&  \multicolumn{2}{l}{the transformation function}\\	
			
			$\overline{\mathbf{z}}_j$&  \multicolumn{2}{l}{the embedding of a given sample $x_j$ in an orthogonal feature space}\\
			
			$\overline{\mathbf{z}}_{j,q}$&  \multicolumn{2}{l}{the $q$-th component of $\mathbf{z}_j$ in the frequency domain}\\
			
			$Q$&  \multicolumn{2}{l}{the total number of frequency components}\\      
			
			$\gamma_q$&  \multicolumn{2}{l}{the weight on the regularization term of approximating the $q$-th frequency component of the old embedding space}\\     
			
			$\Psi(\cdot,\cdot)$&  \multicolumn{2}{l}{the composition function (e.g., for a naive implementation, a simple concatenation operation)}\\ 
			
			$\mathbf{\tilde{z}}_j$&  \multicolumn{2}{l}{the composite feature of sample $x_j$}\\ 
			\bottomrule		
		\end{tabular}%
	}
	
	\label{Notation}%
\end{table*}%

\subsection{Frequency domain learning.} 
Recently, a series of research works explore introducing frequency transformation into deep learning to provide auxiliary information~\cite{su2020collaborative, wang2020towards}. 
Some of these frequency-aware methods aim to reduce the computing cost with frequency transformation~\cite{gueguen2018faster, xu2020learning}, thus improving the network efficiency.
Others propose to conduct frequency-aware augmentation~\cite{kim2020regularization, yang2020fda} to improve the robustness or to solve the problem of domain adaptation.
Apart from that, there is also a work that decouples the entangled information and regularizes the model on some specific components~\cite{khorramshahi2020devil} (e.g., focus more on details in vehicles recognition task).  
Our intra-space SvF analysis is more inspired by the work on frequency-based decouple~\cite{khorramshahi2020devil}, discussing different properties of different frequency components and proposing a frequency-aware regularization method for FSCIL.

%% file: sections_for_review/methods.tex

\section{Methods}
\begin{figure*}[t]
	
	\begin{minipage}[ht]{1\textwidth}
		\centering
		\includegraphics[width = 1\columnwidth]{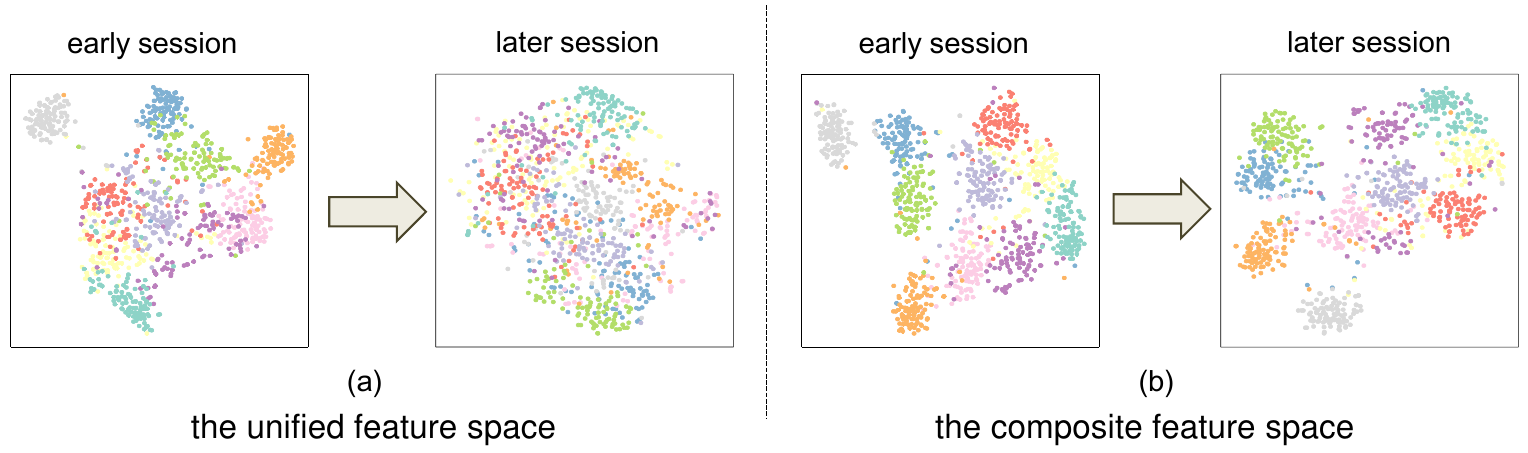}
	\end{minipage}	
	
	\caption{(Best viewed in color.) Visualization of samples in ``the unified feature space'' or ``the composite feature space'' by t-SNE on CIFAR100. Samples of ten classes are from two tasks and each class is represented by one color. (a): Samples in the unified feature space at an early session and a later session; (b): Samples in the composite feature space at an early session and a later session.}
	\label{fig:dual_spaces}
\end{figure*}
\subsection{Few-Shot Class-Incremental Learning}\label{section_two_dot_one}
To better understand our representations, we provide detailed explanations of the main notations and symbols used throughout this paper as shown in Table~\ref{Notation}.

In a few-shot class-incremental learning setup, a network learns several tasks continually, and each task contains a batch of new classes~\cite{tao2020few,rajasegaran2019random,yu2020semantic}. 
The time interval from the arrival of the current task to that of the next task is considered as a FSCIL session~\cite{kemker2017fearnet}.
We suppose the training set for the $t$-th task is $D^{(t)}$ and each $D^{(t)}$ only contains a few samples, except the first task (termed as a base task) $D^{(1)}$, which has a large number of training samples and classes instead. 
$C^{(t)}$ is the set of classes of the $t$-th task, and we consider the generally studied case where there is no overlap between the classes of different tasks: for $ i\neq j, C^{(i)}\cap C^{(j)}=\varnothing$. 
At the $t$-th session, we only have access to the data $D^{(t)}$, and the model is trained on it.

FSCIL can be formulated in two frameworks~\cite{rebuffi2017icarl,yu2020semantic}.
The first framework is composed of a feature extractor and a softmax classifier, and they are trained jointly. The other one only needs to train an embedding network, mapping samples to a feature space where distance represents the semantic discrepancy between samples~\cite{bromley1994signature}, and utilizes an nearest class mean (NCM) classifier for classification~\cite{rebuffi2017icarl,yu2020semantic, mensink2013distance}, which is defined as:
\begin{equation}
\hat{y}_j=\mathop{\mathrm{argmin}}\limits_{c\in \bigcup_i C^{(i)}} {\rm{dist}}(\mathbf{z}_j,\mathbf{u}_c),
\end{equation}
where ${\rm{dist}}(\cdot ,\cdot)$ is the distance metric (e.g., Euclidean distance) and we denote the embedding of a given sample $x_j$ as $\mathbf{z}_j$ and $\mathbf{u}_c$ is the mean of embedding of class $c$~\cite{rebuffi2017icarl}. 

We follow the latter framework in this paper and thus our goal is to obtain a discriminative feature space $\mathbf{z}_j$ performing well on all the seen tasks, which means balancing old knowledge slow-forgetting and new knowledge fast-adapting well at each session. 
In FSCIL, It becomes rather more difficult because the feature space tends to overfit the very few number of samples, and suffers from the degradation of discriminability and generalization ability, as well as catastrophic forgetting.
Therefore, a method needs to be proposed to disentangle the learned knowledge embedded in $\mathbf{z}_j$, determining which knowledge components to be updated slowly or fast, and thereby achieving slow old-knowledge forgetting and fast new-knowledge adaptation.

Our multi-grained ``slow vs. fast" (SvF) few-shot class-incremental learning pipeline (i.e., how to train our embedding model) is detailed in the rest of this section.
An overview of our whole pipeline is shown in Figure~\ref{fig:our_scheme}.
The dash lines in Figure~\ref{fig:our_scheme} illustrate the intra-space grained frequency regularization, which is analyzed in Section~\ref{section_two_dot_two}, while the solid lines show inter-space grained space composition operation, elaborated in Section~\ref{section_two_dot_three}.

\subsection{Intra-Space Level SvF learning}\label{section_two_dot_two}
To train our embedding model, we utilize a metric learning loss to ensure that the distance between similar instances is close, and vice versa. 
The objective function is formulated as:
\begin{equation}
\mathcal{L}=\mathcal{L}_{ML}+\lambda\mathcal{L}_R,
\end{equation}
where $\mathcal{L}_{ML}$ is the metric loss term and $\mathcal{L}_R$ is the regularization term for retaining past knowledge~\cite{li2017learning,kirkpatrick2017overcoming,aljundi2018memory,yu2020semantic}, $\lambda$ denotes the trade-off.
The triplet loss~\cite{wang2014learning} is often adopted as the metric learning loss $\mathcal{L}_{ML}$:
\begin{equation}
\mathcal{L}_{ML}=\max(0,d_+-d_-+r),
\end{equation}
where $d_+$ and $d_-$ are the Euclidean distances between the embeddings of the anchor $x_a$ and the positive instance $x_p$ and the negative instance $x_n$ respectively, and $r$ denotes the margin.
For regularization term $\mathcal{L}_R$, previous methods usually retain old knowledge by distilling on a unified feature space (i.e., directly approximating the embedding $\mathbf{z}^{(t-1)}_j$ while updating $\mathbf{z}^{(t)}_j$  at the $t$-th session):
\begin{equation}\label{regularization_term}
\mathcal{L}_R = \left\|\mathbf{z}^{(t)}_j-\mathbf{z}^{(t-1)}_j\right\|,
\end{equation}
where $\left\|\cdot\right\|$ denotes the Frobenius norm.

To better analyze intra-space SvF learning in an orthogonal space with uncorrelated components, we propose an intra-space SvF feature disentanglement scheme.   
We utilize $T(\cdot)$ to denote the transformation function, and the embedding in the original entangled feature space $\mathbf{z}_j$ is transformed to the embedding in an orthogonal feature space $\overline{\mathbf{z}}_j$:
\begin{equation}
\overline{\mathbf{z}}_j = T(\mathbf{z}_j),
\end{equation} 

In this paper, we utilize Discrete Cosine Transform (DCT) to conduct the transformation. 
In this way, the transformed feature $\overline{\mathbf{z}}_j$ is with the same length as $\mathbf{z}_j$: 
\begin{equation}
\overline{\mathbf{z}}_j = [\overline{\mathbf{z}}_{j,1},\overline{\mathbf{z}}_{j,2},\cdots,\overline{\mathbf{z}}_{j,Q}],
\end{equation}
where $Q$ denotes the total number of frequency components (also the length of $\mathbf{z}_j$), and each component $\overline{\mathbf{z}}_{j,q}$ denotes the $q$-th component of $\mathbf{z}_j$ in the frequency domain. 

As shown in Figure~\ref{fig:frequency}, we find that different frequency components indeed appear to have different characteristics for knowledge transfer when separately distilling them in the fashion of Equation~\ref{regularization_term} to preserve old knowledge.
Specifically, we find that the low-frequency components of the feature space often contribute more to preserving old-knowledge. Therefore, we turn out to approximate the old feature space by frequency-aware knowledge distillation, formulated as:
\begin{equation}\label{frequency_regularization_term}
\mathcal{L}_R = \sum_{q=1}^{Q} \gamma_q\left\|\overline{\mathbf{z}}^{(t)}_{j,q}-\overline{\mathbf{z}}^{(t-1)}_{j,q}\right\|,
\end{equation}
where $\gamma_q$ denotes the weight on the regularization term of approximating the $q$-th frequency component of the old embedding space. To obtain a slow-updated space for old-knowledge preservation, we enforce higher $\gamma_q$ on the low-frequency components in the regularization term for less forgetting and vice versa.

\subsection{Inter-Space Level SvF Learning}\label{section_two_dot_three}
Apart from the above intra-space SvF, we here introduce our inter-space SvF in detail.
Previous works strive to maintain a uniform feature space that balances the old-knowledge slow-forgetting and new-knowledge fast-adapting.
However, since the data of old tasks is non-available and the number of new samples is few, a unified feature space is prone to overfit the new tasks, and the discriminability and generalization ability of the feature space easily degrades.
As shown in Figure~\ref{fig:dual_spaces}~(a), upon the arrival of new tasks, the samples of different classes which are separated at an early session, turn out to overlap with each other at later sessions. 
It indicates that the model suffers catastrophic forgetting after several sessions, and the samples which can be discriminated well at an early session are indistinguishable at the later session. 

To this end, we propose to maintain another separate feature space and set up an inter-space SvF scheme to update the two feature spaces in different fashions.
While the one updated in a fast fashion (e.g., training with a larger learning rate) is prone to new knowledge adaptation, the slowly updated one can better preserve the old knowledge throughout the learning process.

For classification, we propose a feature space composition operation to compose the above two spaces.  
After feature space composition, we can obtain a discriminative feature space, as shown in Figure~\ref{fig:dual_spaces}~(b), where samples are clustered according to their classes and can be well separated.
We use $\Psi(\cdot,\cdot)$ to denote the composition function (e.g., for a naive implementation, a simple concatenation operation).
The composite feature $\mathbf{\tilde{z}}_j$ for sample $x_j$ is denoted as $\mathbf{\tilde{z}}_j=\Psi(\mathbf{z}^{slow}_{j},\mathbf{z}^{fast}_{j})$,
where $\mathbf{z}^{slow}_j$ denotes the embedding in the slow-updated feature space and $\mathbf{z}^{fast}_j$ denotes that in the fast-updated feature space (trained at the current session).
We conduct classification in this composite feature space, which is defined as:
\begin{equation}
\label{eqn:final_pred}
\hat{y}_j=\mathop{\mathrm{argmin}}\limits_{c\in \bigcup_i C^{(i)}} (\mathbf{\tilde{z}}_j-\mathbf{\tilde{u}}_c)^\top \mathbf{A} (\mathbf{\tilde{z}}_j-\mathbf{\tilde{u}}_c),
\end{equation}
\begin{equation}
\mathbf{\tilde{u}}_c=\frac{1}{n_c}\sum_j[y_j=c]\cdot \Psi(\mathbf{u}^{slow}_j,\mathbf{u}^{fast}_j).
\end{equation}
where $\mathbf{A}$ is a metric matrix.
For a simple formulation, $\mathbf{A}$ can be a diagonal matrix and thereby indicating the importance of the slow-updated features and fast-updated features, and all features will be concerned equally if $\mathbf{A}$ is an identity matrix.

%% file: sections_for_review/eval.tex
 \begin{table*}[!ht]
	\centering
	\caption{Comparison results on CUB200 with ResNet18 using the $10$-way $5$-shot FSCIL setting. Our method outperforms others at all learning sessions.}
	\resizebox{1\textwidth}{!}{
		\begin{tabular}{lccccccccccccc}
			\toprule
			\multirow{2}{*}{Method} &\multicolumn{11}{c}{learning sessions}& our relative\\
			\cline{2-12}
			&1 &2 & 3 & 4 & 5 & 6 &7 &8 &9& 10& 11&improvements\\				
			\midrule
			iCaRL-CNN~\cite{rebuffi2017icarl}& 68.68& 52.65& 48.61 & 44.16 & 36.62 & 29.52 &27.83 &26.26 &24.01& 23.89& 21.16&\textbf{+33.17}\\
			EEIL~\cite{castro2018end}& 68.68& 53.63& 47.91 & 44.20 & 36.30 & 27.46 &25.93 &24.70 &23.95& 24.13& 22.11&\textbf{+32.22}\\
			LUCIR-CNN~\cite{hou2019learning}& 68.68& 57.12& 44.21 & 28.78 & 26.71 & 25.66 &24.62 &21.52 &20.12& 20.06& 19.87&\textbf{+34.46}\\
			TOPIC~\cite{tao2020few}& 68.68& 62.49& 54.81 & 49.99 & 45.25 & 41.40 &38.35 &35.36 &32.22& 28.31& 26.28&\textbf{+28.05}\\
			\midrule
			iCaRL-NCM~\cite{rebuffi2017icarl}&72.29&56.67&50.76&43.29&40.99&34.07&30.01&28.83&26.56&23.76&23.32&\textbf{+31.01}\\
			LUCIR-NCM~\cite{hou2019learning}&72.29&58.70&50.68&49.82&45.59&43.10&34.77&31.35&28.53&25.73&22.91&\textbf{+31.42}\\
			SDC~\cite{yu2020semantic}& 72.29&  68.22 & 61.94 & 61.32& 59.83 & 57.30 & 55.48&54.20 &49.99& 48.85&42.58& \textbf{+11.75}\\
			POD~\cite{douillard2020small}& 72.29&  59.77  & 51.23& 48.78 & 47.83 & 44.22&39.76 &37.79& 35.23&31.92&31.27 &\textbf{+23.06}\\
			\midrule
			Ours& 72.29& \textbf{70.53}& \textbf{67.00} & \textbf{64.92} & \textbf{62.67}&  \textbf{61.89}& \textbf{59.63} & \textbf{59.15} & \textbf{57.73}&\textbf{55.92} & \textbf{54.33}\\
			\bottomrule						
		\end{tabular}
	}
	\label{tab:cub:state_few}
\end{table*}

\section{Experiments and Results}

\subsection{Datasets}
CIFAR100~\cite{krizhevsky2009learning} is a labeled subset of the $80$ million tiny image dataset for object recognition. It contains $60000$ $32\times 32$ RGB images in $100$ classes, with $500$ images per class for training and $100$ images per class for testing. CUB200-2011~\cite{wah2011caltech} contains $6000$ images for training and $6000$ for testing with the resolution of $256\times 256$, over $200$ bird categories. It is originally designed for fine-grained image classification.
MiniImageNet is the subset of ImageNet-1k~\cite{vinyals2016matching} that is utilized by few-shot learning. It contains $100$ classes. Each class contains $500$ $84\times 84$ images for training and $100$ images for testing. 
\subsection{Implementation Details}
\subsubsection{Evaluation Protocol}
We conduct experiments under the FSCIL setting.
Following~\cite{tao2020few}, we evaluate our method on three datasets (CIFAR100, MiniImageNet, and CUB200) with similar evaluation protocols.
For each $D^{(t)}, t\textgreater 1$, if the number of classes $|C^{(t)}|$ is $M$ and the number of training samples per class is $K$, we denote the setting as $M$-way $K$-shot. For CIFAR100 and miniImageNet datasets, we choose $60$ and $40$ classes for the base task and new tasks, respectively, and adopt the $5$-way $5$-shot setting, leading to $9$ training sessions in total. For CUB200, we adopt the $10$-way $5$-shot setting, by picking $100$ classes into $10$ new learning sessions and the base task has $100$ classes. For all datasets, we construct the training set of each learning session by randomly selecting $5$ training samples per class from the original dataset, and the test set is the same as the original one. After training on a new batch of classes, we evaluate the trained model on test samples of all seen classes.

\begin{figure*}[t]
	\begin{minipage}[ht]{1\textwidth}
		\centering
		\includegraphics[width =0.33\columnwidth]{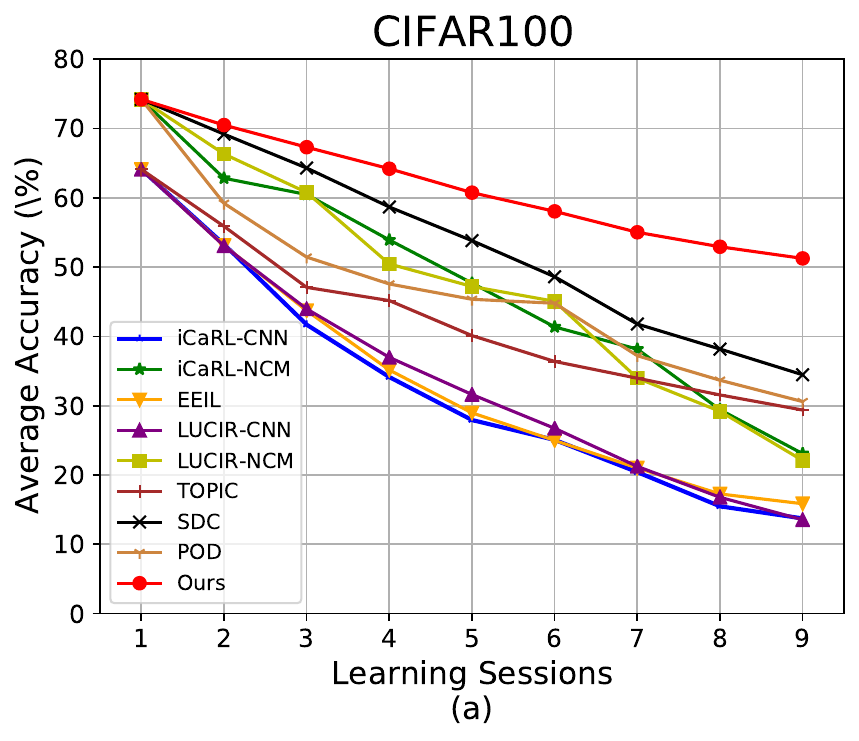}	
		\includegraphics[width = 0.33\columnwidth]{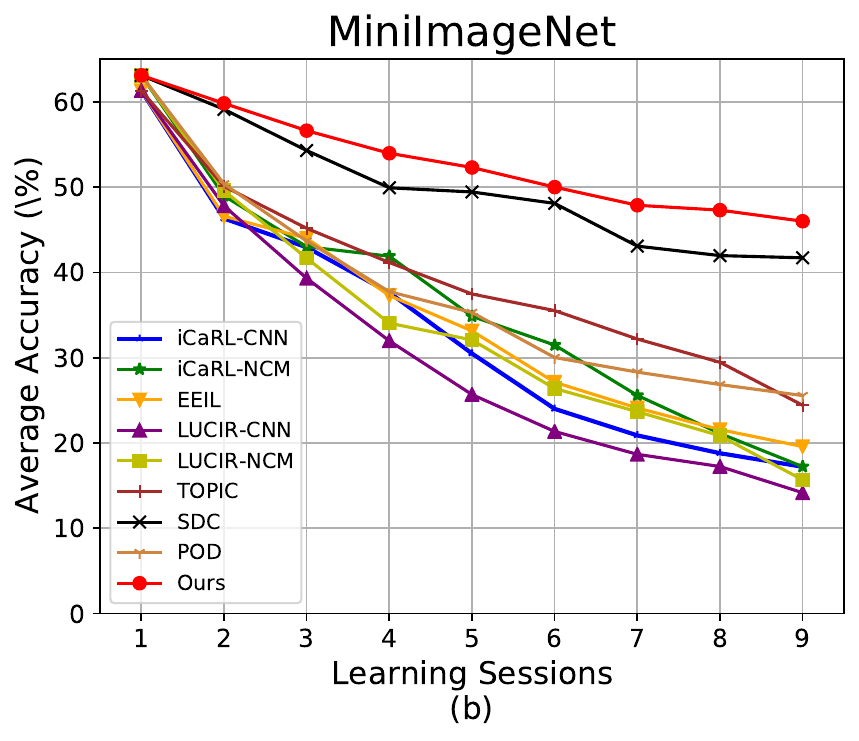}
		\includegraphics[width = 0.33\columnwidth]{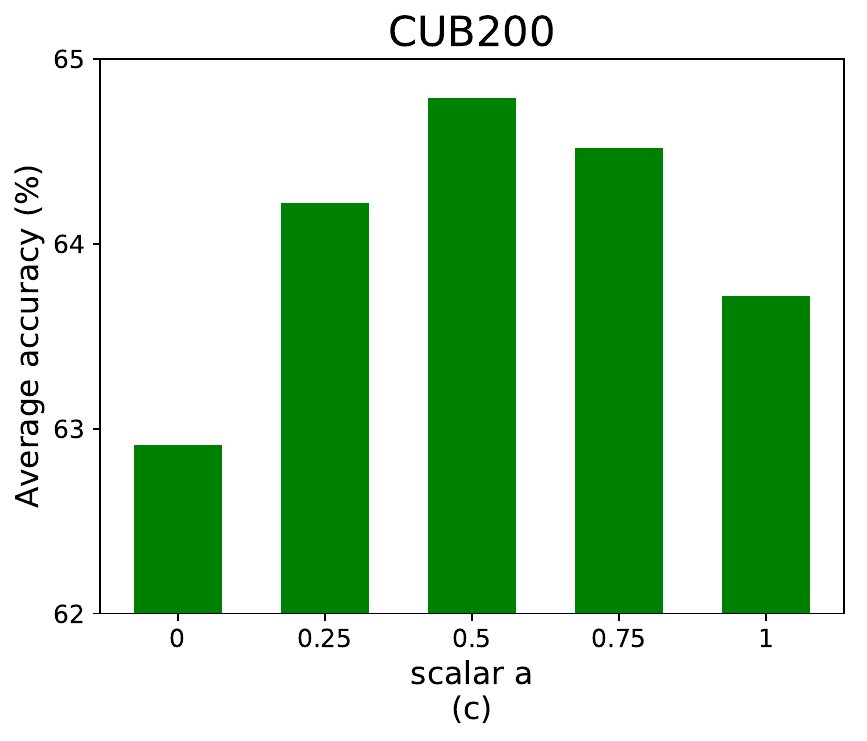}
	\end{minipage}	
	\caption{
		(a): Comparison results on CIFAR100 with ResNet18 using the $5$-way $5$-shot FSCIL setting. 
		(b): Comparison results on MiniImageNet with ResNet18 using the $5$-way $5$-shot FSCIL setting.
		Our method shows clear superiority and outperforms all other methods at each encountered learning session.
		(c): Change of average accuracy when varying $a$ on CUB200.The performance peaks on an intermediate value, which indicates the importance and complementarity of both the slow-updated space and the fast-updated space.}
	\label{fig:cifar:state_few}
\end{figure*}

\begin{figure*}
	\begin{minipage}[ht]{1\textwidth}
		\centering
		\includegraphics[width = 0.33\columnwidth]{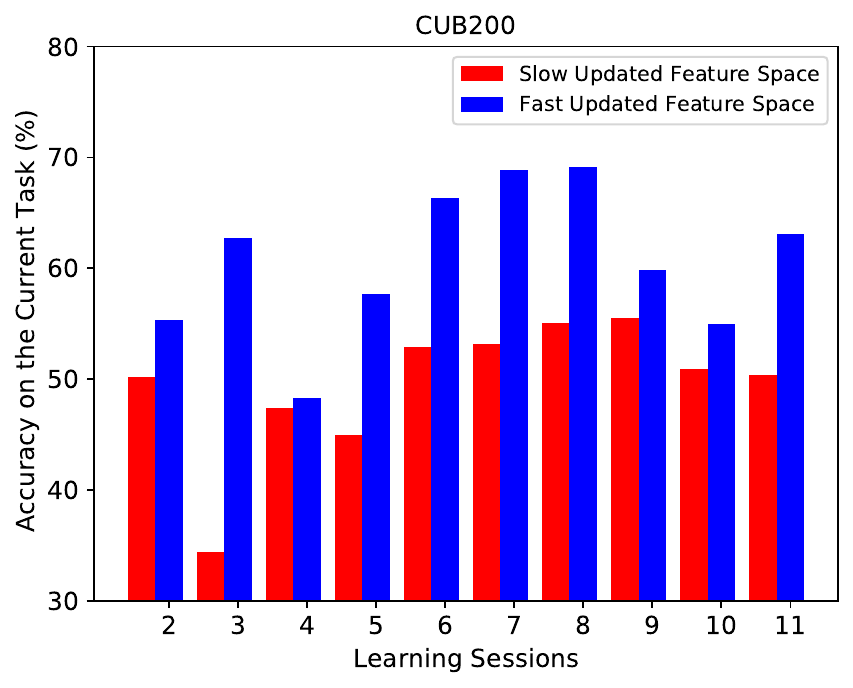}
		\includegraphics[width = 0.33\columnwidth]{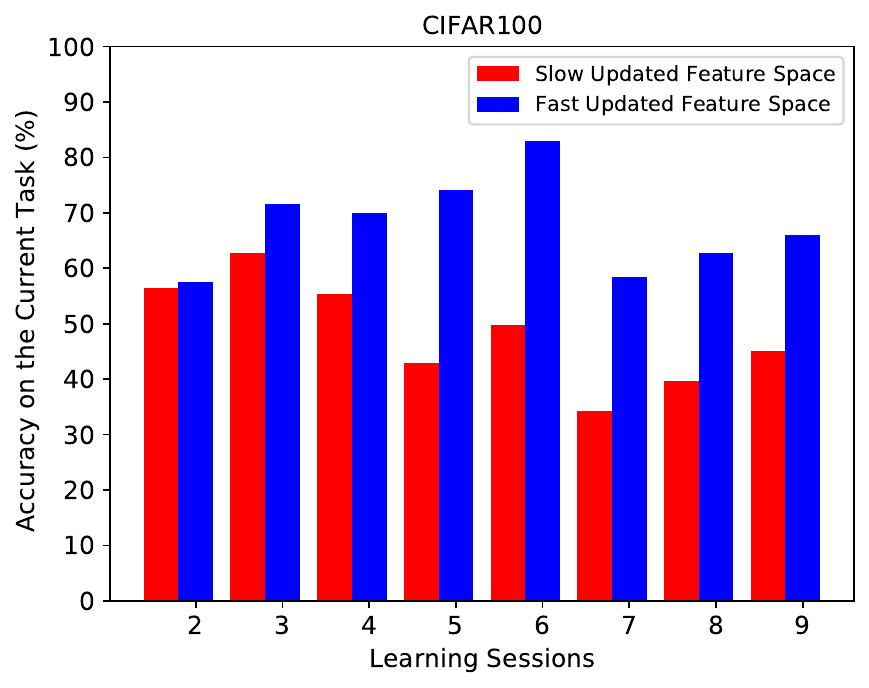}
		\includegraphics[width = 0.33\columnwidth]{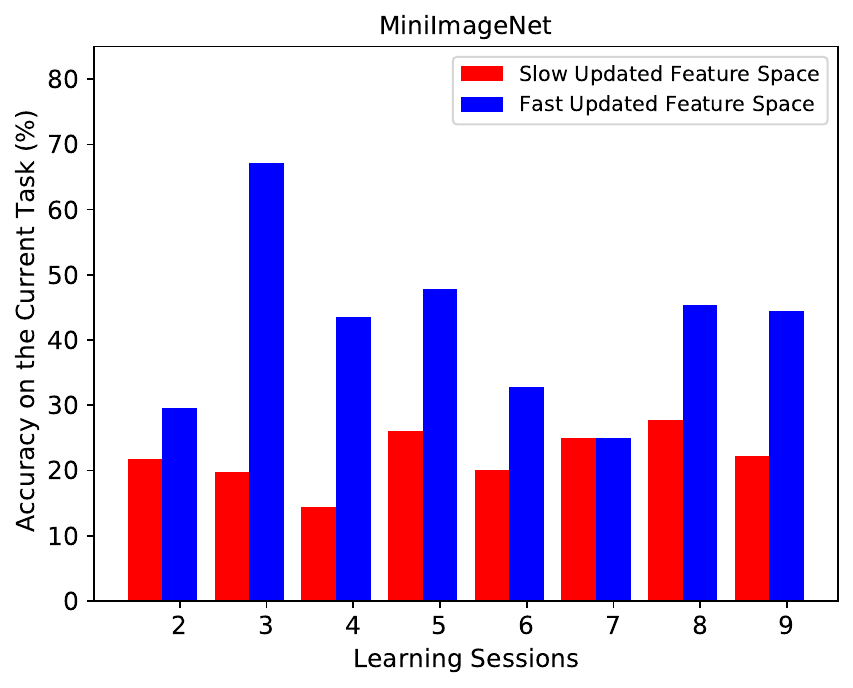}
		\caption{Accuracy for the current task while only utilizing the slow-updated feature space or the fast-updated feature space on CUB200, CIFAR100, and MiniImageNet. A large performance gap on the data of the current task exists between the slow-updated space and fast-updated space.} 
		\label{fig:base+life-long:ablation}	
	\end{minipage}		
\end{figure*}

\begin{figure*}[!h]
	\centering
	\begin{minipage}[ht]{0.24\textwidth}
		\includegraphics[width = 1\columnwidth]{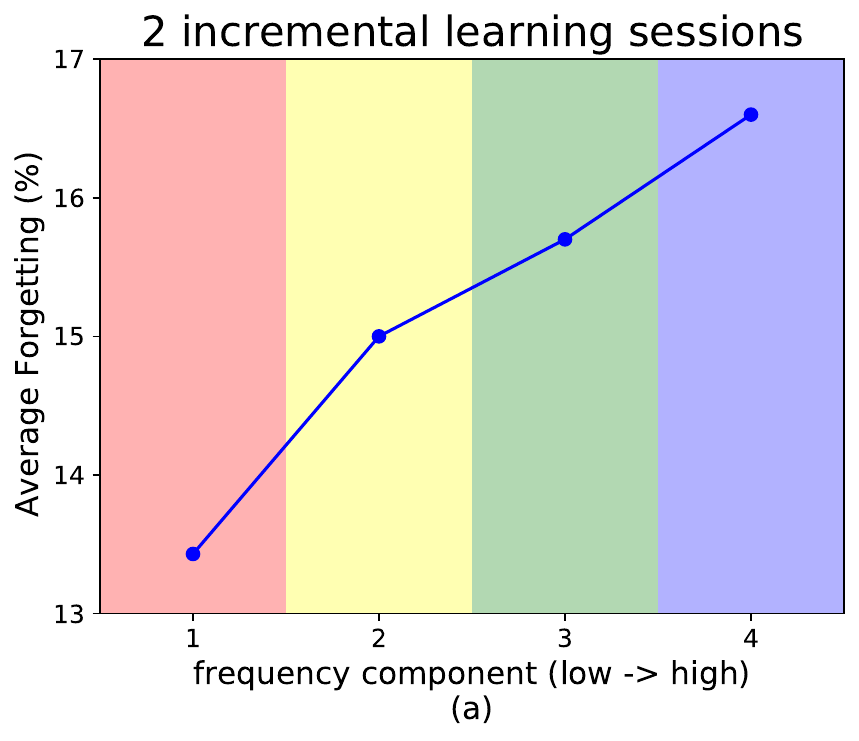}
	\end{minipage}	
	\begin{minipage}[ht]{0.24\textwidth}	
		\includegraphics[width = 1\columnwidth]{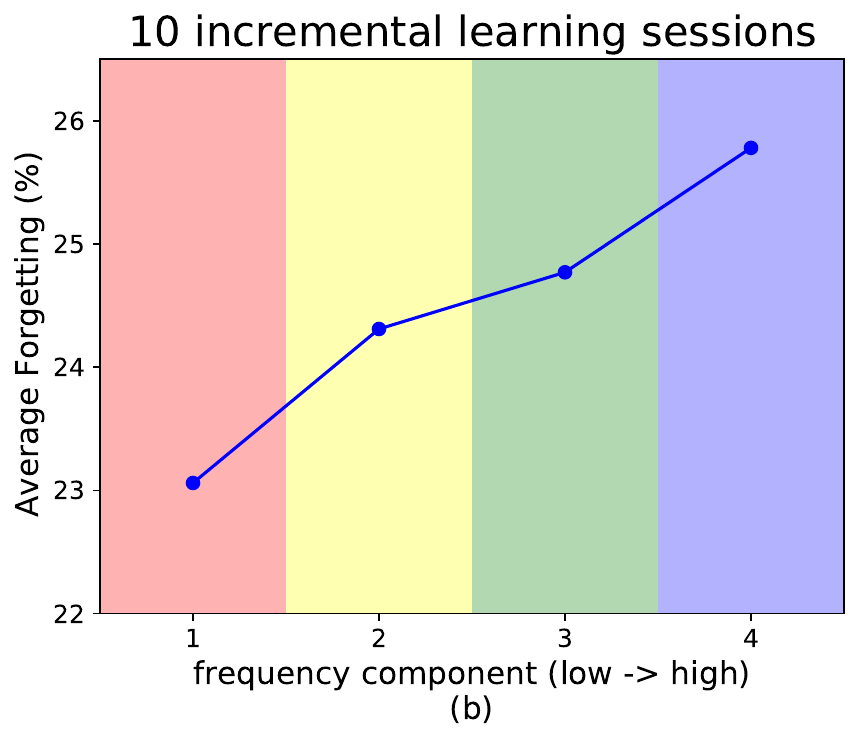}
	\end{minipage}
	\begin{minipage}[ht]{0.24\textwidth}
		\includegraphics[width = 1\columnwidth]{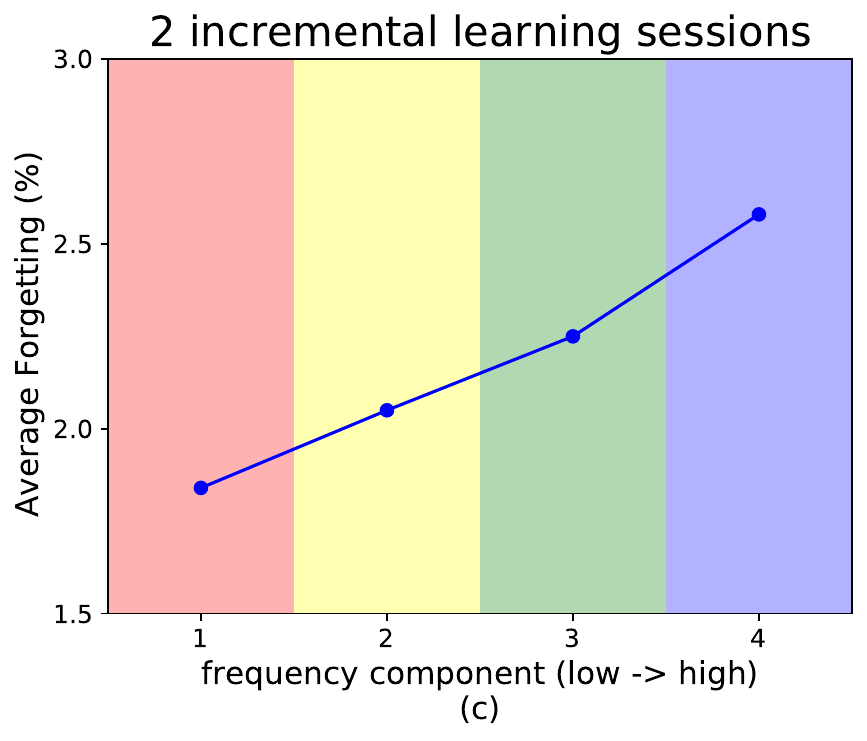}
	\end{minipage}	
	\begin{minipage}[ht]{0.24\textwidth}	
		\includegraphics[width = 1\columnwidth]{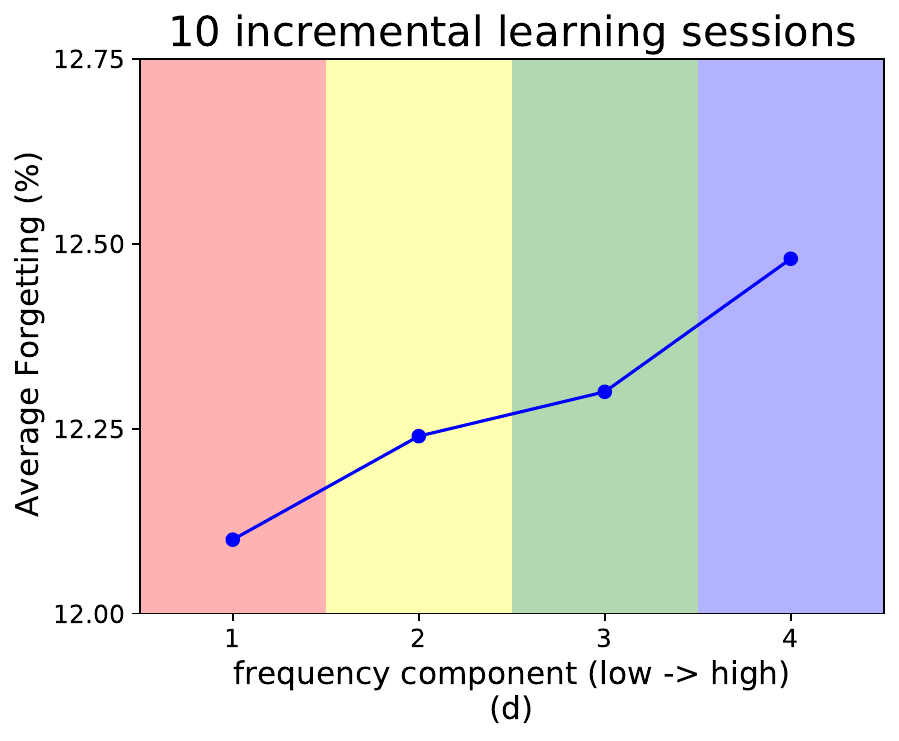}
	\end{minipage}
	\caption{Analysis of intra-space SvF.
		(a) Results with 2 learning sessions on CUB200. (b) Results with 10 learning sessions on CUB200. (c) Results with 2 learning sessions on MiniImageNet. (d) Results with 10 learning sessions on MiniImageNet. 
	}
	\label{sfig:frequency:cub+mini}
\end{figure*}

\subsubsection{Training Details} We implement our models with Pytorch and use Adam~\cite{kingma2014adam} for optimization. Following~\cite{tao2020few}, we use the ResNet18~\cite{he2016deep} as the backbone network. The model is trained with the same strategy in~\cite{tao2020few} for a fair comparison. 
We use an embedding network as the feature extractor and the dimensions of the feature extractor are $512$. The base model is trained with the same strategy in~\cite{tao2020few}. 
For inter-space SvF learning, we train our slow-updated models with learning rate $1e-6$ for 50 epochs and train our fast-updated models with learning rate $1e-5$ for 50 epochs. 
We follow the setting discussed in the part ``Inter-space SvF analysis'' of Section~\ref{section:ablation} and choose the scalar $a=0.5$ as default for the feature space composition operation.
For intra-space SvF learning, we obtain $512$ frequency components from the original feature by DCT, and the overall spectrum is divided into $8$ groups. The DCT operation is implemented with torch-dct project on github. The time for training a model on MiniImageNet only increases by around $2\%$ (memory $4\%$), which are almost negligible. 
We conduct frequency-aware regularization by adjusting the weights of these $8$ frequency groups (i.e. $\{\gamma_q\}^{8}_{q=1}$).
For the slow-updated models, when updating on each new task, we set the weight $\gamma_q=1$ if $q = 1$ otherwise $\gamma_q=0$. For the fast-updated models, we set the weight $\gamma_q=0$ if $q = 1$ otherwise $\gamma_q=1$. We compute the centers of old classes after training one task and fix them in following tasks, which is standard as SDC~\cite{yu2020semantic}.
For data augmentation, we use standard random cropping and flipping as in~\cite{tao2020few}. 


\subsection{Comparison to State-of-the-Art Methods}
In this section, we compare our proposed method in the few-shot class-incremental learning scenario, with existing state-of-the-art methods. They include iCaRL~\cite{rebuffi2017icarl}, EEIL~\cite{castro2018end}, LUCIR~\cite{hou2019learning}, TOPIC~\cite{tao2020few}, SDC~\cite{yu2020semantic}, POD~\cite{douillard2020small}. We report the results of iCaRL and LUCIR with both a softmax classifier and an NCM classifier, which are denoted as iCaRL-CNN and iCaRL-NCM (as well as LUCIR-CNN and LUCIR-NCM) respectively. Using different classifiers achieves different accuracy at the first task (the base task).
Table~\ref{tab:cub:state_few} reports the test accuracy of different methods on CUB-200 dataset. 
Compared with the others, our method shows clear superiority (more than $10\%$). As shown in Figure~\ref{fig:cifar:state_few} (a) and Figure~\ref{fig:cifar:state_few} (b), we can observe that our method outperforms all other methods at each encountered learning session on CIFAR100 and MiniImageNet datasets.



\begin{table}[t]
	\centering
	\caption{Validation of our intra-space frequency-aware knowledge distillation and inter-space composition operation on CUB200, CIFAR100 and MiniImageNet. The performance at the last learning session (i.e., Last) and the average results over all the learning sessions (i.e., Average) are reported here.}
	\resizebox{0.499\textwidth}{!}{
		\begin{tabular}{lcccccc}
			\toprule
			\multirow{2}{*}{Method}&\multicolumn{2}{c}{CUB200}	&\multicolumn{2}{c}{CIFAR100}&\multicolumn{2}{c}{MiniImageNet}	\\
			\cline{2-7}
			&Last&Average&Last&Average&Last&Average\\
			\midrule
			Baseline &42.58&57.45&34.46&53.68&41.71&49.96\\
			\midrule
			Baseline+intra-space&49.96&59.70&\textbf{50.60}&\textbf{61.09}&44.46&\textbf{51.93}\\
			Baseline+inter-space &\textbf{53.16}&\textbf{60.68}&49.11&60.41&\textbf{44.72}&51.90\\
			\bottomrule
		\end{tabular}
	}
	\label{tab:inter_space:ablation}
\end{table}

\subsection{Ablation Study}
\label{section:ablation}
In this section, we carry out ablation experiments to validate our inter-space grain and intra-space grain SvF strategies.
Extensive experiments are conducted to show the effect of training samples' number, feature space composition method, etc.
We also explore the properties of fast-updated space, slow-updated space, as well as different frequency components when transferring old knowledge and further elaborated on the inter-space and intra-space SvF analysis, respectively.  

\subsubsection{Baseline} As described in Section~\ref{section_two_dot_one}, we follow SDC~\cite{yu2020semantic} and implement our method with an embedding network and an NCM classifier. We adopt the triplet loss~\cite{wang2014learning} as the metric loss, and conduct knowledge distillation in the embedding space~\cite{li2017learning} to retain old knowledge.

\subsubsection{Effect of Intra-space and Inter-space SvF Strategies} 
We first conduct ablation experiments to show the effectiveness of our inter-space and intra-space SvF strategies, respectively.
As shown in Table~\ref{tab:inter_space:ablation}, both the inter-space grain and intra-space grain SvF strategies improve the performance of our baseline.
Inter-space SvF strategy achieves better performance than the intra-space SvF strategy on CUB200 (around 1\%) which only contains the samples of bird categories. On the datasets which contain more diverse classes (e.g., CIFAR100 and MiniImageNet), the intra-space SvF strategy achieves comparable performance to the inter-space SvF strategy.
For inter-space SvF, we also evaluate the performance of only using the slow-updated feature space alone for classification.
As shown in Figure~\ref{fig:base+life-long:ablation}, 
a large performance gap on the data of the current task exists between the slow-updated feature space and the fast-updated one, which shows the importance of the fast-updated space to fit new knowledge.
It indicates the reasonability of inter-space SvF. In this way, it is difficult to keep a good balance between slow-forgetting for old knowledge and fast-adapting for new knowledge by adjusting the updating fashion of a unified feature space.

\begin{table}[t]
	\centering
	\caption{Results with different inter-space composition operations on CUB200, CIFAR100, and MiniImageNet. The sophisticated composition operation implemented with PCA outperforms the simple version by around $1\%$.}
	\resizebox{0.499\textwidth}{!}{
		\begin{tabular}{lcccccc}
			\toprule
			\multirow{2}{*}{Method}&\multicolumn{2}{c}{CUB200}	&\multicolumn{2}{c}{CIFAR100}&\multicolumn{2}{c}{MiniImageNet}	\\
			\cline{2-7}
			&Last &Average&Last&Average&Last&Average\\
			\midrule
			Baseline &42.58&57.45&34.46&53.68&41.71&49.96\\
			\midrule
			Baseline+inter-space-simple &53.16&60.68&49.11&60.41&44.72&51.90\\
			Baseline+inter-space-pca&\textbf{53.76}&\textbf{61.15}&\textbf{50.51}&\textbf{60.78}&\textbf{45.38}&\textbf{53.24}\\
			\bottomrule
		\end{tabular}
	}
	\label{tab:class_PCA:ablation}
\end{table}

\subsubsection{Intra-space SvF Analysis}  
We here introduce our intra-space SvF analysis, examining the effect of different frequency components for old-knowledge preservation.
We divide the overall spectrum into four groups and regularize the model with one of them separately.
It can be observed in Figure~\ref{fig:frequency} that different groups appear different characteristics for old-knowledge transferring, and a pretty clear trend can be found that the forgetting rate of old tasks increases along with the frequency.
It indicates that the regularization of low-frequency components is more conducive to preserving old knowledge than that on high-frequency components.
Therefore, we enforce higher weights on those low-frequency components in the regularization term to make these knowledge components of the feature space updating more slowly. The results on CUB200 and MiniImageNet are shown in Figure~\ref{sfig:frequency:cub+mini}.


\subsubsection{Inter-Space SvF analysis}
We here briefly show that the fast-updated space and the slow-updated space are complementary, even with the simplest implementation strategy for space composition, where $\mathbf{A}$ is constructed as $\mathbf{A}=\left[\begin{array}{cc}a\mathbf{I}& 0 \\ 0 & (1-a)\mathbf{I}\end{array}\right]$ with a scalar $a$ ( $\mathbf{I}$ is an identity matrix with dimension half of $\mathbf{A}$'s).
$a=0$ means that only using the slow-updated feature space and $a=1$ for only using the fast-updated feature space at the current session.
The change of accuracy, with respect to $a$, is shown in Figure~\ref{fig:cifar:state_few}(c). Using a slow-updated feature space achieves the lowest accuracy since it contains limited new-task knowledge. The performance of using a fast-updated feature space independently is also lower than that of the composite space, because of forgetting old knowledge.
The performance peaks on an intermediate value, which indicates the complementarity.
More sophisticated forms of the metric matrix $\mathbf{A}$ can also be constructed (e.g., in data-driven learning), and the discussion and analysis of another space composition strategy are detailed in the next paragraph.

\begin{figure*}[t]
	
	\begin{minipage}[t]{1\textwidth}
		\centering
		\includegraphics[width = 0.33\columnwidth]{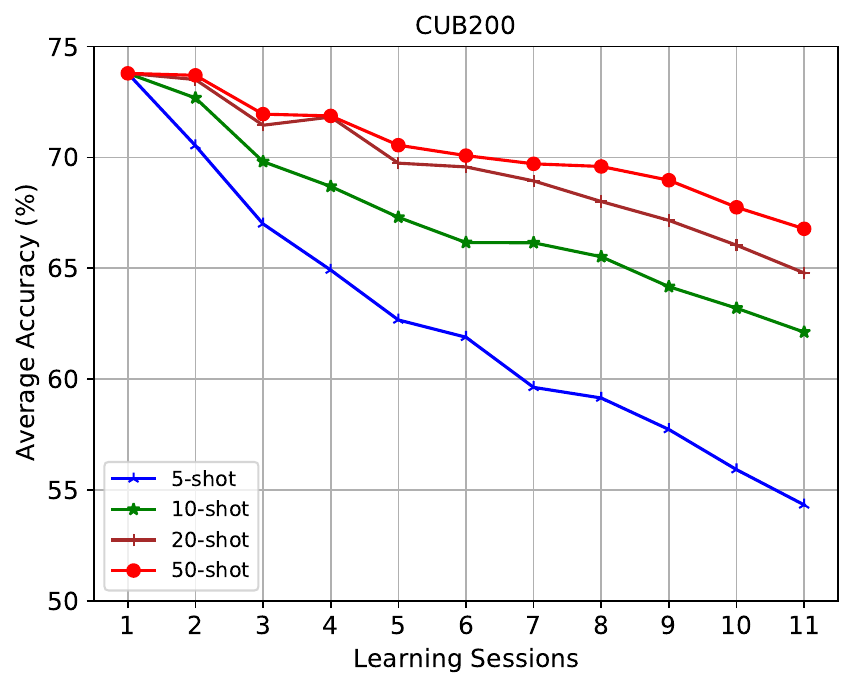}
		\includegraphics[width = 0.33\columnwidth]{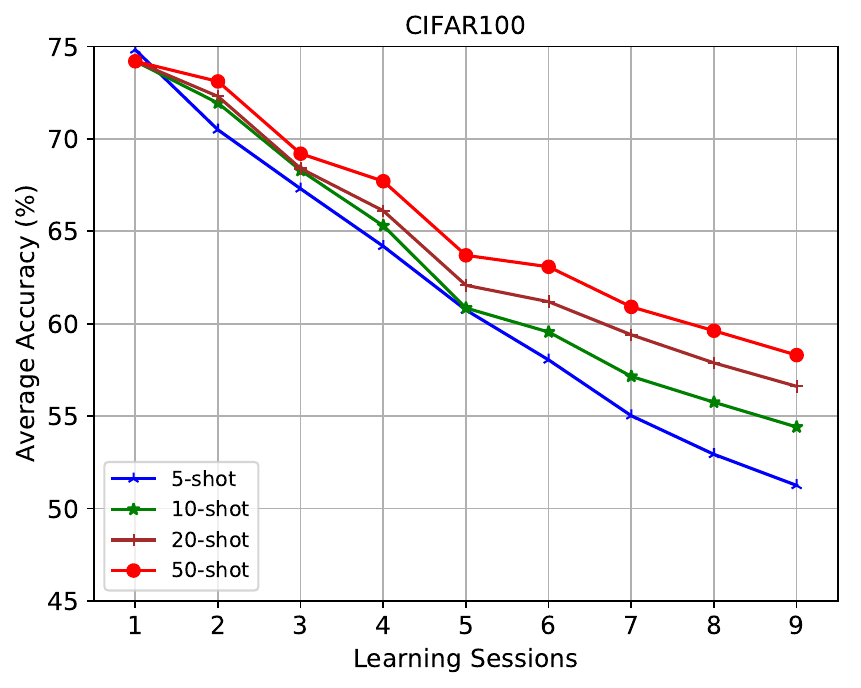}
		\includegraphics[width = 0.33\columnwidth]{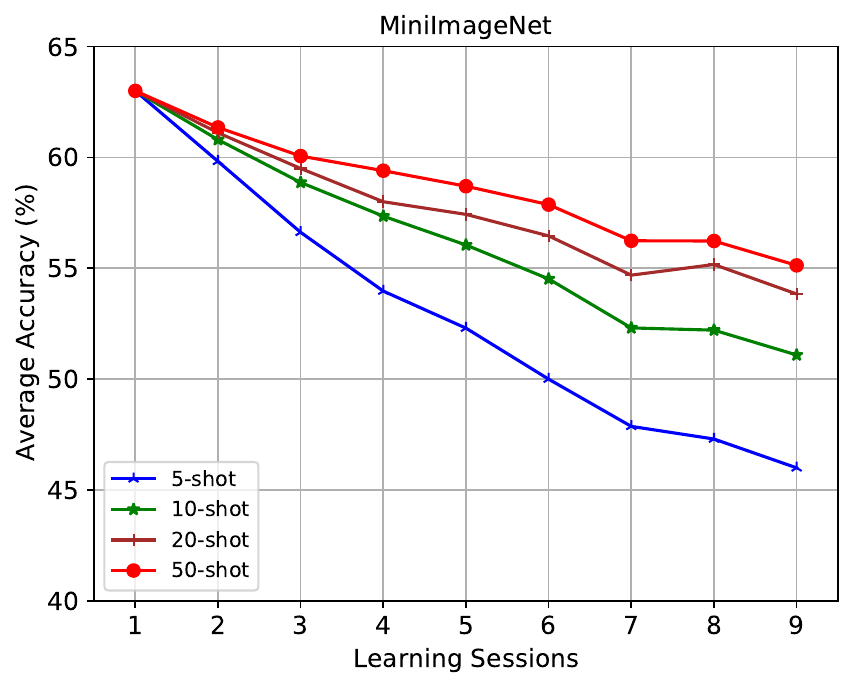}
		\caption{Results on CUB200, CIFAR100, and MiniImageNet under different FSCIL settings (5-shot, 10-shot, 20-shot, and 50-shot). The performance of our method increases as the number of training samples increases.}
		\label{fig:different_few_shot:ablation}
	\end{minipage}			
\end{figure*}

\begin{figure*}[t]
	\begin{minipage}[t]{1\textwidth}
		\centering
		\includegraphics[width = 0.33\columnwidth]{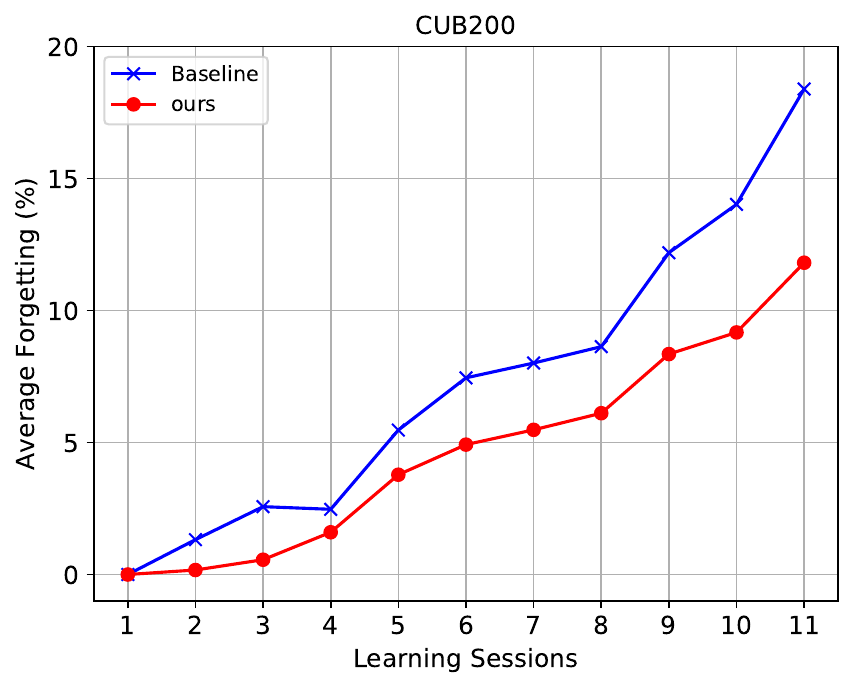}
		\includegraphics[width = 0.33\columnwidth]{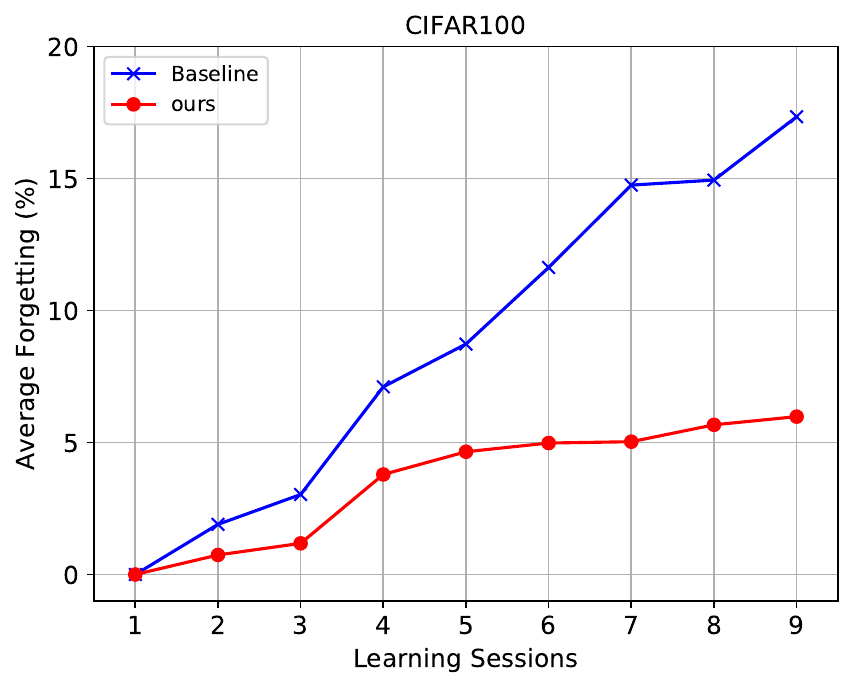}
		\includegraphics[width = 0.33\columnwidth]{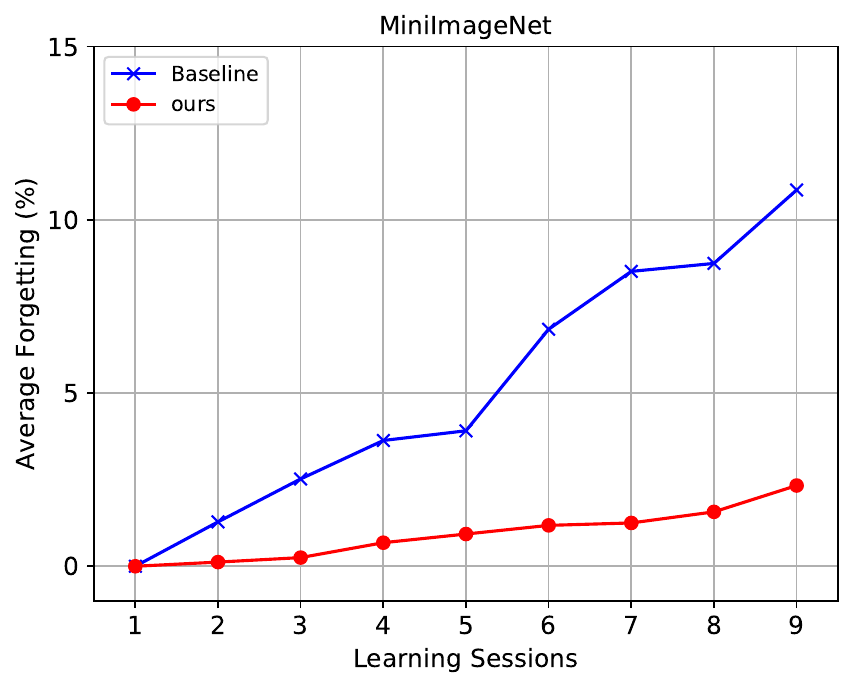}
		\caption{Average forgetting of our method and the baseline with a unified feature space on CUB200, CIFAR100, and MiniImageNet. Our method outperforms the baseline by a large margin at the last learning session.}
		\label{fig:cub_forgetting:ablation}	
	\end{minipage}
\end{figure*}
\subsubsection{Different Inter-space Composition Operations} 
We evaluate another more sophisticated inter-space composition method which first reduces the dimensions of the features from the fast-updated space and slow-updated space by principal component analysis (PCA), and then concatenates them.
The results are shown in Table~\ref{tab:class_PCA:ablation}, where the simplest composition strategy is denoted by ``inter-space-simple'', and the sophisticated one is denoted by ``inter-space-pca''. 
Specifically, we compute the PCA transformation matrix $\mathbf{P}_1$ and $\mathbf{P}_2$ with embeddings of samples obtained by the slowly updated model and the fast updated model respectively. Another choice is learning $\mathbf{P}_1$ and $\mathbf{P}_2$ incrementally~\cite{hall1998incremental,weng2003candid} with samples from the subsequent tasks. 
Then we reduce the dimension to, e.g., half of the original, and therefore the size of $\mathbf{P}_1$ and $\mathbf{P}_2$ are both $256 \times 512$.
For concatenating features in the composite space, the transformation matrix can be constructed as $ \mathbf{Q}=\left[\begin{array}{cc}\mathbf{P}_1& 0 \\ 0 & \mathbf{P}_2\end{array}\right]$.
Given a feature $\mathbf{\tilde{z}}_j$, and a center $\mathbf{\tilde{u}}_c$ for class $c$, the dimension-reduced data can be $\mathbf{Q}\mathbf{\tilde{z}}_j$ and $\mathbf{Q}\mathbf{\tilde{u}}_c$, respectively.
And the classification can be formulated as:
\begin{equation}
\begin{aligned}
\hat{y}_j&=\mathop{\mathrm{argmin}}\limits_{c\in \bigcup_i C^{(i)}} (\mathbf{Q}\mathbf{\tilde{z}}_j-\mathbf{Q}\mathbf{\tilde{u}}_c)^\top (\mathbf{Q}\mathbf{\tilde{z}}_j-\mathbf{Q}\mathbf{\tilde{u}}_c)\\
&=\mathop{\mathrm{argmin}}\limits_{c\in \bigcup_i C^{(i)}} (\mathbf{\tilde{z}}_j-\mathbf{\tilde{u}}_c)^\top \mathbf{Q}^\top\mathbf{Q} (\mathbf{\tilde{z}}_j-\mathbf{\tilde{u}}_c).
\end{aligned}
\end{equation}
Compared with Equation~\eqref{eqn:final_pred}, matrix $\mathbf{Q}^\top\mathbf{Q}$ can be considered as a specific data-driven metric matrix $\mathbf{A}$.
Also, the transformation matrix $\mathbf{Q}$ can be viewed as a part of composition function $\Psi(\cdot,\cdot)$. We set $\mathbf{P}_1 = \mathbf{P}_2$ in the few-shot scenario because such few new-task samples are not able to estimate a reasonable $\mathbf{P}_2$.  As shown in Table~\ref{tab:class_PCA:ablation}, ``Baseline+inter-space-pca'' achieves $1\%$ higher accuracy. 

\subsubsection{The Effect of The Number of Training Samples} 
To examine the effect of the number of training samples, we evaluate our method with different shots of training samples, which are $5$-shot, $10$-shot, $20$-shot, and $50$-shot settings. As shown in Figure~\ref{fig:different_few_shot:ablation}, we can see that the performance of our method increases as the number of training samples increases. 
It can also be noticed that the number of samples matters more on latter training sessions, and the performance gap grows more rapidly when the number of samples gets larger.

\subsubsection{Average Forgetting of Previous Tasks} The forgetting of previous tasks is estimated with \textit{average forgetting}~\cite{yu2020semantic,chaudhry2018riemannian}.
We illustrate the forgetting curves of our method across $11$ learning sessions on CUB200 ($9$ learning sessions on CIFAR100 and MiniImageNet), shown in Figure~\ref{fig:cub_forgetting:ablation}. On these three datasets,
we can observe that our method outperforms ``Baseline'' by a large margin at the last learning session (more than $5\%$). These results indicate the stability of our composite feature space against the continuous arrival of new tasks.

\subsubsection{Analysis of How to Choose Hyperparameters}
The hyperparameters (i.e. gammas and learning rates) are chosen according to the following rules: 1) for gammas (e.g. $\{\gamma_q\}^{8}_{q=1}$ correspond to $8$ frequency groups from low-frequency to high-frequency), we set higher weights $\gamma_i=1$ on low-frequency groups (the first group) and lower weights $\gamma_j=0$ on high-frequency groups (the other groups) when training slow-updated model, and vice versa for fast-updated one.
Considering the first group as the low-frequency group usually leads to best performance. 2) for learning rates, the learning rate of fast-updated model is usually \textbf{$10$} times than that of slow-updated one. In our experiments, we choose 1e-5 and 1e-6 for fast-updated and slow-updated model respectively.

We have also evaluated our method with different numbers of frequency groups (i.e. $N_Q=2, 4, 8, 16$) on MiniImageNet. The average results over all learning sessions are 52.76\%, 53.03\%, 53.24\%, 52.78\% respectively. Our method is stable as the number of frequency groups changes and achieves the best performance when $N_Q=8$.  In our experiments, we choose $N_Q=8$ as the number of frequency groups.

%% file: sections_for_review/conclusion.tex
\section{Conclusion}
In this paper, we propose a novel few-shot class-incremental learning scheme based on a ``slow vs. fast" (SvF) learning pipeline. 
In the pipeline, we design an intra-space frequency-aware regularization to 
enforce a SvF constraint on different frequency components, and present an inter-space composition operation to well balance the slow forgetting of old knowledge and fast adaptation to new knowledge. Comprehensive experimental results demonstrate that the proposed approach significantly outperforms other state-of-the-art approaches by a large margin. 

%% file: tpami.bbl
\begin{thebibliography}{10}
\providecommand{\url}[1]{#1}
\csname url@samestyle\endcsname
\providecommand{\newblock}{\relax}
\providecommand{\bibinfo}[2]{#2}
\providecommand{\BIBentrySTDinterwordspacing}{\spaceskip=0pt\relax}
\providecommand{\BIBentryALTinterwordstretchfactor}{4}
\providecommand{\BIBentryALTinterwordspacing}{\spaceskip=\fontdimen2\font plus
\BIBentryALTinterwordstretchfactor\fontdimen3\font minus
  \fontdimen4\font\relax}
\providecommand{\BIBforeignlanguage}[2]{{%
\expandafter\ifx\csname l@#1\endcsname\relax
\typeout{** WARNING: IEEEtran.bst: No hyphenation pattern has been}%
\typeout{** loaded for the language `#1'. Using the pattern for}%
\typeout{** the default language instead.}%
\else
\language=\csname l@#1\endcsname
\fi
#2}}
\providecommand{\BIBdecl}{\relax}
\BIBdecl

\bibitem{yu2020semantic}
L.~Yu, B.~Twardowski, X.~Liu, L.~Herranz, K.~Wang, Y.~Cheng, S.~Jui, and
  J.~van~de Weijer, ``Semantic drift compensation for class-incremental
  learning,'' in \emph{Proceedings of the IEEE conference on computer vision
  and pattern recognition (CVPR)}, 2020.

\bibitem{tao2020topology}
X.~Tao, X.~Chang, X.~Hong, X.~Wei, and Y.~Gong, ``Topology-preserving
  class-incremental learning.''\hskip 1em plus 0.5em minus 0.4em\relax ECCV,
  2020.

\bibitem{iscen2020memory}
A.~Iscen, J.~Zhang, S.~Lazebnik, and C.~Schmid, ``Memory-efficient incremental
  learning through feature adaptation.''\hskip 1em plus 0.5em minus 0.4em\relax
  ECCV, 2020.

\bibitem{douillard2020small}
A.~Douillard, M.~Cord, C.~Ollion, T.~Robert, and E.~Valle, ``Small-task
  incremental learning.''\hskip 1em plus 0.5em minus 0.4em\relax ECCV, 2020.

\bibitem{zhao2020maintaining}
B.~Zhao, X.~Xiao, G.~Gan, B.~Zhang, and S.-T. Xia, ``Maintaining discrimination
  and fairness in class incremental learning,'' in \emph{Proceedings of the
  IEEE/CVF Conference on Computer Vision and Pattern Recognition}, 2020, pp.
  13\,208--13\,217.

\bibitem{rajasegaran2020itaml}
J.~Rajasegaran, S.~Khan, M.~Hayat, F.~S. Khan, and M.~Shah, ``itaml: An
  incremental task-agnostic meta-learning approach,'' in \emph{Proceedings of
  the IEEE/CVF Conference on Computer Vision and Pattern Recognition}, 2020,
  pp. 13\,588--13\,597.

\bibitem{hayes2019remind}
T.~L. Hayes, K.~Kafle, R.~Shrestha, M.~Acharya, and C.~Kanan, ``Remind your
  neural network to prevent catastrophic forgetting.''\hskip 1em plus 0.5em
  minus 0.4em\relax ECCV, 2020.

\bibitem{masana2020class}
M.~Masana, X.~Liu, B.~Twardowski, M.~Menta, A.~D. Bagdanov, and J.~van~de
  Weijer, ``Class-incremental learning: survey and performance evaluation,''
  \emph{arXiv preprint arXiv:2010.15277}, 2020.

\bibitem{tao2020few}
X.~Tao, X.~Hong, X.~Chang, S.~Dong, X.~Wei, and Y.~Gong, ``Few-shot
  class-incremental learning,'' in \emph{Proceedings of the IEEE conference on
  computer vision and pattern recognition (CVPR)}, 2020.

\bibitem{chaudhry2018riemannian}
A.~Chaudhry, P.~K. Dokania, T.~Ajanthan, and P.~H. Torr, ``Riemannian walk for
  incremental learning: Understanding forgetting and intransigence,'' in
  \emph{Proceedings of the European Conference on Computer Vision (ECCV)},
  2018, pp. 532--547.

\bibitem{french1999catastrophic}
R.~M. French, ``Catastrophic forgetting in connectionist networks,''
  \emph{Trends in cognitive sciences}, vol.~3, no.~4, pp. 128--135, 1999.

\bibitem{goodfellow2013empirical}
I.~J. Goodfellow, M.~Mirza, D.~Xiao, A.~Courville, and Y.~Bengio, ``An
  empirical investigation of catastrophic forgetting in gradient-based neural
  networks,'' in \emph{Proceedings of the International Conference on Learning
  Representations (ICLR)}, 2014.

\bibitem{mccloskey1989catastrophic}
M.~McCloskey and N.~J. Cohen, ``Catastrophic interference in connectionist
  networks: The sequential learning problem,'' in \emph{Psychology of learning
  and motivation}.\hskip 1em plus 0.5em minus 0.4em\relax Elsevier, 1989,
  vol.~24, pp. 109--165.

\bibitem{pfulb2019comprehensive}
B.~Pf{\"u}lb and A.~Gepperth, ``A comprehensive, application-oriented study of
  catastrophic forgetting in dnns,'' in \emph{Proceedings of the International
  Conference on Learning Representations (ICLR)}, 2019.

\bibitem{rebuffi2017icarl}
S.-A. Rebuffi, A.~Kolesnikov, G.~Sperl, and C.~H. Lampert, ``icarl: Incremental
  classifier and representation learning,'' in \emph{Proceedings of the IEEE
  conference on computer vision and pattern recognition (CVPR)}, 2017.

\bibitem{castro2018end}
F.~M. Castro, M.~J. Mar{\'\i}n-Jim{\'e}nez, N.~Guil, C.~Schmid, and K.~Alahari,
  ``End-to-end incremental learning,'' in \emph{Proceedings of the European
  Conference on Computer Vision (ECCV)}, 2018, pp. 233--248.

\bibitem{hou2019learning}
S.~Hou, X.~Pan, C.~C. Loy, Z.~Wang, and D.~Lin, ``Learning a unified classifier
  incrementally via rebalancing,'' in \emph{Proceedings of the IEEE conference
  on computer vision and pattern recognition (CVPR)}, 2019.

\bibitem{de2019continual}
M.~De~Lange, R.~Aljundi, M.~Masana, S.~Parisot, X.~Jia, A.~Leonardis,
  G.~Slabaugh, and T.~Tuytelaars, ``A continual learning survey: Defying
  forgetting in classification tasks,'' \emph{arXiv preprint arXiv:1909.08383},
  2019.

\bibitem{parisi2019continual}
G.~I. Parisi, R.~Kemker, J.~L. Part, C.~Kanan, and S.~Wermter, ``Continual
  lifelong learning with neural networks: A review,'' \emph{Neural Networks},
  2019.

\bibitem{li2020compositional}
Y.~Li, L.~Zhao, K.~Church, and M.~Elhoseiny, ``Compositional continual language
  learning,'' in \emph{Proceedings of the International Conference on Learning
  Representations (ICLR)}, 2020.

\bibitem{lee2020neural}
S.~Lee, J.~Ha, D.~Zhang, and G.~Kim, ``A neural dirichlet process mixture model
  for task-free continual learning,'' in \emph{Proceedings of the International
  Conference on Learning Representations (ICLR)}, 2020.

\bibitem{adel2019continual}
T.~Adel, H.~Zhao, and R.~E. Turner, ``Continual learning with adaptive weights
  (claw),'' in \emph{Proceedings of the International Conference on Learning
  Representations (ICLR)}, 2020.

\bibitem{von2019continual}
J.~von Oswald, C.~Henning, J.~Sacramento, and B.~F. Grewe, ``Continual learning
  with hypernetworks,'' in \emph{Proceedings of the International Conference on
  Learning Representations (ICLR)}, 2020.

\bibitem{kurlecontinual}
R.~Kurle, B.~Cseke, A.~Klushyn, P.~van~der Smagt, and S.~G{\"u}nnemann,
  ``Continual learning with bayesian neural networks for non-stationary data,''
  in \emph{Proceedings of the International Conference on Learning
  Representations (ICLR)}, 2020.

\bibitem{titsias2019functional}
M.~K. Titsias, J.~Schwarz, A.~G. d.~G. Matthews, R.~Pascanu, and Y.~W. Teh,
  ``Functional regularisation for continual learning with gaussian processes,''
  in \emph{Proceedings of the International Conference on Learning
  Representations (ICLR)}, 2019.

\bibitem{ebrahimi2019uncertainty}
S.~Ebrahimi, M.~Elhoseiny, T.~Darrell, and M.~Rohrbach, ``Uncertainty-guided
  continual learning with bayesian neural networks,'' in \emph{Proceedings of
  the International Conference on Learning Representations (ICLR)}, 2020.

\bibitem{zeng2019continual}
G.~Zeng, Y.~Chen, B.~Cui, and S.~Yu, ``Continual learning of context-dependent
  processing in neural networks,'' \emph{Nature Machine Intelligence}, vol.~1,
  no.~8, pp. 364--372, 2019.

\bibitem{kundu2020class}
J.~N. Kundu, R.~M. Venkatesh, N.~Venkat, A.~Revanur, and R.~V. Babu,
  ``Class-incremental domain adaptation,'' 2020.

\bibitem{ostapenko2019learning}
O.~Ostapenko, M.~Puscas, T.~Klein, P.~Jahnichen, and M.~Nabi, ``Learning to
  remember: A synaptic plasticity driven framework for continual learning,'' in
  \emph{Proceedings of the IEEE Conference on Computer Vision and Pattern
  Recognition}, 2019, pp. 11\,321--11\,329.

\bibitem{aljundi2019task}
R.~Aljundi, K.~Kelchtermans, and T.~Tuytelaars, ``Task-free continual
  learning,'' in \emph{Proceedings of the IEEE Conference on Computer Vision
  and Pattern Recognition}, 2019, pp. 11\,254--11\,263.

\bibitem{lee2020continual}
J.~Lee, H.~G. Hong, D.~Joo, and J.~Kim, ``Continual learning with extended
  kronecker-factored approximate curvature,'' in \emph{Proceedings of the
  IEEE/CVF Conference on Computer Vision and Pattern Recognition}, 2020, pp.
  9001--9010.

\bibitem{he2020incremental}
J.~He, R.~Mao, Z.~Shao, and F.~Zhu, ``Incremental learning in online
  scenario,'' in \emph{Proceedings of the IEEE/CVF Conference on Computer
  Vision and Pattern Recognition}, 2020, pp. 13\,926--13\,935.

\bibitem{ebrahimi2020adversarial}
S.~Ebrahimi, F.~Meier, R.~Calandra, T.~Darrell, and M.~Rohrbach, ``Adversarial
  continual learning.''\hskip 1em plus 0.5em minus 0.4em\relax ECCV, 2020.

\bibitem{liumore}
Y.~Liu, S.~Parisot, G.~Slabaugh, X.~Jia, A.~Leonardis, and T.~Tuytelaars,
  ``More classifiers, less forgetting: A generic multi-classifier paradigm for
  incremental learning.''\hskip 1em plus 0.5em minus 0.4em\relax ECCV, 2020.

\bibitem{caccia2020online}
M.~Caccia, P.~Rodriguez, O.~Ostapenko, F.~Normandin, M.~Lin, L.~Page-Caccia,
  I.~H. Laradji, I.~Rish, A.~Lacoste, D.~V{\'a}zquez \emph{et~al.}, ``Online
  fast adaptation and knowledge accumulation (osaka): a new approach to
  continual learning,'' \emph{Advances in Neural Information Processing
  Systems}, vol.~33, 2020.

\bibitem{nguyen2017variational}
C.~V. Nguyen, Y.~Li, T.~D. Bui, and R.~E. Turner, ``Variational continual
  learning,'' \emph{arXiv preprint arXiv:1710.10628}, 2017.

\bibitem{chen2020incremental}
J.~Chen, S.~Wang, L.~Chen, H.~Cai, and Y.~Qian, ``Incremental detection of
  remote sensing objects with feature pyramid and knowledge distillation,''
  \emph{IEEE Transactions on Geoscience and Remote Sensing}, 2020.

\bibitem{dang2020class}
S.~Dang, Z.~Cao, Z.~Cui, Y.~Pi, and N.~Liu, ``Class boundary exemplar selection
  based incremental learning for automatic target recognition,'' \emph{IEEE
  Transactions on Geoscience and Remote Sensing}, vol.~58, no.~8, pp.
  5782--5792, 2020.

\bibitem{shim2020online}
D.~Shim, Z.~Mai, J.~Jeong, S.~Sanner, H.~Kim, and J.~Jang, ``Online
  class-incremental continual learning with adversarial shapley value,''
  \emph{arXiv e-prints}, pp. arXiv--2009, 2020.

\bibitem{liu2020incdet}
L.~Liu, Z.~Kuang, Y.~Chen, J.-H. Xue, W.~Yang, and W.~Zhang, ``Incdet: in
  defense of elastic weight consolidation for incremental object detection,''
  \emph{IEEE transactions on neural networks and learning systems}, 2020.

\bibitem{li2020continual}
H.~Li, P.~Barnaghi, S.~Enshaeifar, and F.~Ganz, ``Continual learning using
  bayesian neural networks,'' \emph{IEEE Transactions on Neural Networks and
  Learning Systems}, 2020.

\bibitem{li2011graph}
X.~Li, A.~Dick, H.~Wang, C.~Shen, and A.~van~den Hengel, ``Graph mode-based
  contextual kernels for robust svm tracking,'' in \emph{2011 international
  conference on computer vision}.\hskip 1em plus 0.5em minus 0.4em\relax IEEE,
  2011, pp. 1156--1163.

\bibitem{chen2014ranking}
Y.~Chen, X.~Li, A.~Dick, and R.~Hill, ``Ranking consistency for image matching
  and object retrieval,'' \emph{Pattern Recognition}, vol.~47, no.~3, pp.
  1349--1360, 2014.

\bibitem{jiang2019learning}
X.~Jiang, L.~Zhang, P.~Lv, Y.~Guo, R.~Zhu, Y.~Li, Y.~Pang, X.~Li, B.~Zhou, and
  M.~Xu, ``Learning multi-level density maps for crowd counting,'' \emph{IEEE
  transactions on neural networks and learning systems}, vol.~31, no.~8, pp.
  2705--2715, 2019.

\bibitem{yoon2017lifelong}
J.~Yoon, E.~Yang, J.~Lee, and S.~J. Hwang, ``Lifelong learning with dynamically
  expandable networks,'' in \emph{Proceedings of the International Conference
  on Learning Representations (ICLR)}, 2018.

\bibitem{li2019learn}
X.~Li, Y.~Zhou, T.~Wu, R.~Socher, and C.~Xiong, ``Learn to grow: A continual
  structure learning framework for overcoming catastrophic forgetting,''
  \emph{International Conference on Machine Learning (ICML)}, 2019.

\bibitem{hung2019compacting}
C.-Y. Hung, C.-H. Tu, C.-E. Wu, C.-H. Chen, Y.-M. Chan, and C.-S. Chen,
  ``Compacting, picking and growing for unforgetting continual learning,'' in
  \emph{Advances in Neural Information Processing Systems}, 2019, pp.
  13\,647--13\,657.

\bibitem{mallya2018packnet}
A.~Mallya and S.~Lazebnik, ``Packnet: Adding multiple tasks to a single network
  by iterative pruning,'' in \emph{Proceedings of the IEEE conference on
  computer vision and pattern recognition (CVPR)}, 2018.

\bibitem{mallya2018piggyback}
A.~Mallya, D.~Davis, and S.~Lazebnik, ``Piggyback: Adapting a single network to
  multiple tasks by learning to mask weights,'' in \emph{Proceedings of the
  European Conference on Computer Vision (ECCV)}, 2018, pp. 67--82.

\bibitem{serra2018overcoming}
J.~Serr{\`a}, D.~Sur{\'\i}s, M.~Miron, and A.~Karatzoglou, ``Overcoming
  catastrophic forgetting with hard attention to the task,'' \emph{arXiv
  preprint arXiv:1801.01423}, 2018.

\bibitem{rusu2016progressive}
A.~A. Rusu, N.~C. Rabinowitz, G.~Desjardins, H.~Soyer, J.~Kirkpatrick,
  K.~Kavukcuoglu, R.~Pascanu, and R.~Hadsell, ``Progressive neural networks,''
  \emph{arXiv preprint arXiv:1606.04671}, 2016.

\bibitem{aljundi2017expert}
R.~Aljundi, P.~Chakravarty, and T.~Tuytelaars, ``Expert gate: Lifelong learning
  with a network of experts,'' in \emph{Proceedings of the IEEE conference on
  computer vision and pattern recognition (CVPR)}, 2017.

\bibitem{rajasegaran2019random}
J.~Rajasegaran, M.~Hayat, S.~H. Khan, F.~S. Khan, and L.~Shao, ``Random path
  selection for continual learning,'' in \emph{Advances in Neural Information
  Processing Systems}, 2019, pp. 12\,648--12\,658.

\bibitem{abati2020conditional}
D.~Abati, J.~Tomczak, T.~Blankevoort, S.~Calderara, R.~Cucchiara, and B.~E.
  Bejnordi, ``Conditional channel gated networks for task-aware continual
  learning,'' in \emph{Proceedings of the IEEE/CVF Conference on Computer
  Vision and Pattern Recognition}, 2020, pp. 3931--3940.

\bibitem{chaudhry2019agem}
A.~Chaudhry, M.~Ranzato, M.~Rohrbach, and M.~Elhoseiny, ``Efficient lifelong
  learning with a-gem,'' in \emph{Proceedings of the International Conference
  on Learning Representations (ICLR)}, 2019.

\bibitem{lopez2017gradient}
D.~Lopez-Paz and M.~Ranzato, ``Gradient episodic memory for continual
  learning,'' in \emph{Advances in Neural Information Processing Systems},
  2017, pp. 6467--6476.

\bibitem{shin2017continual}
H.~Shin, J.~K. Lee, J.~Kim, and J.~Kim, ``Continual learning with deep
  generative replay,'' in \emph{Advances in Neural Information Processing
  Systems}, 2017, pp. 2990--2999.

\bibitem{aljundi2019online}
R.~Aljundi, E.~Belilovsky, T.~Tuytelaars, L.~Charlin, M.~Caccia, M.~Lin, and
  L.~Page-Caccia, ``Online continual learning with maximal interfered
  retrieval,'' in \emph{Advances in Neural Information Processing Systems},
  2019, pp. 11\,849--11\,860.

\bibitem{zhai2019lifelong}
M.~Zhai, L.~Chen, F.~Tung, J.~He, M.~Nawhal, and G.~Mori, ``Lifelong gan:
  Continual learning for conditional image generation,'' in \emph{Proceedings
  of the IEEE International Conference on Computer Vision}, 2019, pp.
  2759--2768.

\bibitem{wu2018memory}
C.~Wu, L.~Herranz, X.~Liu, J.~van~de Weijer, B.~Raducanu \emph{et~al.},
  ``Memory replay gans: Learning to generate new categories without
  forgetting,'' in \emph{Advances In Neural Information Processing Systems},
  2018, pp. 5962--5972.

\bibitem{he2018exemplar}
C.~He, R.~Wang, S.~Shan, and X.~Chen, ``Exemplar-supported generative
  reproduction for class incremental learning.'' in \emph{BMVC}, 2018, p.~98.

\bibitem{wu2019large}
Y.~Wu, Y.~Chen, L.~Wang, Y.~Ye, Z.~Liu, Y.~Guo, and Y.~Fu, ``Large scale
  incremental learning,'' in \emph{Proceedings of the IEEE conference on
  computer vision and pattern recognition (CVPR)}, 2019.

\bibitem{liu2020mnemonics}
Y.~Liu, A.-A. Liu, Y.~Su, B.~Schiele, and Q.~Sun, ``Mnemonics training:
  Multi-class incremental learning without forgetting,'' \emph{arXiv preprint
  arXiv:2002.10211}, 2020.

\bibitem{summaira2021recent}
J.~Summaira, X.~Li, A.~M. Shoib, S.~Li, and J.~Abdul, ``Recent advances and
  trends in multimodal deep learning: A review,'' \emph{arXiv preprint
  arXiv:2105.11087}, 2021.

\bibitem{lee2017overcoming}
S.-W. Lee, J.-H. Kim, J.~Jun, J.-W. Ha, and B.-T. Zhang, ``Overcoming
  catastrophic forgetting by incremental moment matching,'' in \emph{Advances
  in Neural Information Processing Systems}, 2017, pp. 4652--4662.

\bibitem{li2017learning}
Z.~Li and D.~Hoiem, ``Learning without forgetting,'' \emph{IEEE transactions on
  pattern analysis and machine intelligence}, vol.~40, no.~12, pp. 2935--2947,
  2017.

\bibitem{ritter2018online}
H.~Ritter, A.~Botev, and D.~Barber, ``Online structured laplace approximations
  for overcoming catastrophic forgetting,'' in \emph{Advances in Neural
  Information Processing Systems}, 2018, pp. 3738--3748.

\bibitem{kirkpatrick2017overcoming}
J.~Kirkpatrick, R.~Pascanu, N.~Rabinowitz, J.~Veness, G.~Desjardins, A.~A.
  Rusu, K.~Milan, J.~Quan, T.~Ramalho, A.~Grabska-Barwinska \emph{et~al.},
  ``Overcoming catastrophic forgetting in neural networks,'' \emph{Proceedings
  of the national academy of sciences}, vol. 114, no.~13, pp. 3521--3526, 2017.

\bibitem{zenke2017continual}
F.~Zenke, B.~Poole, and S.~Ganguli, ``Continual learning through synaptic
  intelligence,'' \emph{International Conference on Machine Learning (ICML)},
  2017.

\bibitem{liu2018rotate}
X.~Liu, M.~Masana, L.~Herranz, J.~Van~de Weijer, A.~M. Lopez, and A.~D.
  Bagdanov, ``Rotate your networks: Better weight consolidation and less
  catastrophic forgetting,'' in \emph{2018 24th International Conference on
  Pattern Recognition (ICPR)}.\hskip 1em plus 0.5em minus 0.4em\relax IEEE,
  2018, pp. 2262--2268.

\bibitem{aljundi2018memory}
R.~Aljundi, F.~Babiloni, M.~Elhoseiny, M.~Rohrbach, and T.~Tuytelaars, ``Memory
  aware synapses: Learning what (not) to forget,'' in \emph{Proceedings of the
  European Conference on Computer Vision (ECCV)}, 2018, pp. 139--154.

\bibitem{dhar2019learning}
P.~Dhar, R.~V. Singh, K.-C. Peng, Z.~Wu, and R.~Chellappa, ``Learning without
  memorizing,'' in \emph{Proceedings of the IEEE conference on computer vision
  and pattern recognition (CVPR)}, 2019.

\bibitem{mirzadeh2020understanding}
S.~I. Mirzadeh, M.~Farajtabar, R.~Pascanu, and H.~Ghasemzadeh, ``Understanding
  the role of training regimes in continual learning,'' \emph{Advances in
  Neural Information Processing Systems}, vol.~33, 2020.

\bibitem{kanakis2020reparameterizing}
M.~Kanakis, D.~Bruggemann, S.~Saha, S.~Georgoulis, A.~Obukhov, and L.~Van~Gool,
  ``Reparameterizing convolutions for incremental multi-task learning without
  task interference,'' \emph{arXiv preprint arXiv:2007.12540}, 2020.

\bibitem{rosenfeld2018incremental}
A.~Rosenfeld and J.~K. Tsotsos, ``Incremental learning through deep
  adaptation,'' \emph{IEEE transactions on pattern analysis and machine
  intelligence}, vol.~42, no.~3, pp. 651--663, 2018.

\bibitem{perez2020incremental}
J.-M. Perez-Rua, X.~Zhu, T.~M. Hospedales, and T.~Xiang, ``Incremental few-shot
  object detection,'' in \emph{Proceedings of the IEEE/CVF Conference on
  Computer Vision and Pattern Recognition}, 2020, pp. 13\,846--13\,855.

\bibitem{cermelli2020modeling}
F.~Cermelli, M.~Mancini, S.~R. Bulo, E.~Ricci, and B.~Caputo, ``Modeling the
  background for incremental learning in semantic segmentation,'' in
  \emph{Proceedings of the IEEE/CVF Conference on Computer Vision and Pattern
  Recognition}, 2020, pp. 9233--9242.

\bibitem{belouadah2018deesil}
E.~Belouadah and A.~Popescu, ``Deesil: Deep-shallow incremental learning.'' in
  \emph{Proceedings of the European Conference on Computer Vision (ECCV)},
  2018, pp. 0--0.

\bibitem{xiang2019incremental}
Y.~Xiang, Y.~Fu, P.~Ji, and H.~Huang, ``Incremental learning using conditional
  adversarial networks,'' in \emph{Proceedings of the IEEE International
  Conference on Computer Vision}, 2019, pp. 6619--6628.

\bibitem{su2020collaborative}
H.~Su, J.~Su, D.~Wang, W.~Gan, W.~Wu, M.~Wang, J.~Yan, and Y.~Qiao,
  ``Collaborative distillation in the parameter and spectrum domains for video
  action recognition,'' \emph{arXiv preprint arXiv:2009.06902}, 2020.

\bibitem{wang2020towards}
Z.~Wang, Y.~Yang, A.~Shrivastava, V.~Rawal, and Z.~Ding, ``Towards
  frequency-based explanation for robust cnn,'' \emph{arXiv preprint
  arXiv:2005.03141}, 2020.

\bibitem{gueguen2018faster}
L.~Gueguen, A.~Sergeev, B.~Kadlec, R.~Liu, and J.~Yosinski, ``Faster neural
  networks straight from jpeg,'' in \emph{Advances in Neural Information
  Processing Systems}, 2018, pp. 3933--3944.

\bibitem{xu2020learning}
K.~Xu, M.~Qin, F.~Sun, Y.~Wang, Y.-K. Chen, and F.~Ren, ``Learning in the
  frequency domain,'' in \emph{Proceedings of the IEEE/CVF Conference on
  Computer Vision and Pattern Recognition}, 2020, pp. 1740--1749.

\bibitem{kim2020regularization}
J.~Kim, S.~Cha, D.~Wee, S.~Bae, and J.~Kim, ``Regularization on
  spatio-temporally smoothed feature for action recognition,'' in
  \emph{Proceedings of the IEEE/CVF Conference on Computer Vision and Pattern
  Recognition}, 2020, pp. 12\,103--12\,112.

\bibitem{yang2020fda}
Y.~Yang and S.~Soatto, ``Fda: Fourier domain adaptation for semantic
  segmentation,'' in \emph{Proceedings of the IEEE/CVF Conference on Computer
  Vision and Pattern Recognition}, 2020, pp. 4085--4095.

\bibitem{khorramshahi2020devil}
P.~Khorramshahi, N.~Peri, J.-c. Chen, and R.~Chellappa, ``The devil is in the
  details: Self-supervised attention for vehicle re-identification,''
  \emph{arXiv preprint arXiv:2004.06271}, 2020.

\bibitem{kemker2017fearnet}
R.~Kemker and C.~Kanan, ``Fearnet: Brain-inspired model for incremental
  learning,'' in \emph{Proceedings of the International Conference on Learning
  Representations (ICLR)}, 2018.

\bibitem{bromley1994signature}
J.~Bromley, I.~Guyon, Y.~LeCun, E.~S{\"a}ckinger, and R.~Shah, ``Signature
  verification using a" siamese" time delay neural network,'' in \emph{Advances
  in neural information processing systems}, 1994, pp. 737--744.

\bibitem{mensink2013distance}
T.~Mensink, J.~Verbeek, F.~Perronnin, and G.~Csurka, ``Distance-based image
  classification: Generalizing to new classes at near-zero cost,'' \emph{IEEE
  transactions on pattern analysis and machine intelligence}, vol.~35, no.~11,
  pp. 2624--2637, 2013.

\bibitem{wang2014learning}
J.~Wang, Y.~Song, T.~Leung, C.~Rosenberg, J.~Wang, J.~Philbin, B.~Chen, and
  Y.~Wu, ``Learning fine-grained image similarity with deep ranking,'' in
  \emph{Proceedings of the IEEE conference on computer vision and pattern
  recognition (CVPR)}, 2014.

\bibitem{krizhevsky2009learning}
A.~Krizhevsky and G.~Hinton, ``Learning multiple layers of features from tiny
  images,'' Citeseer, Tech. Rep., 2009.

\bibitem{wah2011caltech}
C.~Wah, S.~Branson, P.~Welinder, P.~Perona, and S.~Belongie, ``The caltech-ucsd
  birds-200-2011 dataset. technical report cns-tr-2011-001,'' \emph{California
  Institute of Technology}, 2011.

\bibitem{vinyals2016matching}
O.~Vinyals, C.~Blundell, T.~Lillicrap, D.~Wierstra \emph{et~al.}, ``Matching
  networks for one shot learning,'' in \emph{Advances in Neural Information
  Processing Systems}, 2016, pp. 3630--3638.

\bibitem{kingma2014adam}
D.~P. Kingma and J.~Ba, ``Adam: A method for stochastic optimization,'' in
  \emph{Proceedings of the International Conference on Learning Representations
  (ICLR)}, 2014.

\bibitem{he2016deep}
K.~He, X.~Zhang, S.~Ren, and J.~Sun, ``Deep residual learning for image
  recognition,'' in \emph{Proceedings of the IEEE conference on computer vision
  and pattern recognition (CVPR)}, 2016.

\bibitem{hall1998incremental}
P.~M. Hall, A.~D. Marshall, and R.~R. Martin, ``Incremental eigenanalysis for
  classification.'' in \emph{BMVC}, vol.~98.\hskip 1em plus 0.5em minus
  0.4em\relax Citeseer, 1998, pp. 286--295.

\bibitem{weng2003candid}
J.~Weng, Y.~Zhang, and W.-S. Hwang, ``Candid covariance-free incremental
  principal component analysis,'' \emph{IEEE Transactions on Pattern Analysis
  and Machine Intelligence}, vol.~25, no.~8, pp. 1034--1040, 2003.

\end{thebibliography}
